\pdfoutput=1

\documentclass[11pt]{article}

\usepackage[]{acl}

\usepackage{times}
\usepackage{latexsym}

\usepackage[T1]{fontenc}

\usepackage[utf8]{inputenc}

\usepackage{microtype}

\usepackage{inconsolata}

\usepackage{inconsolata}
\usepackage{graphicx}
\usepackage{multirow}
\usepackage{tabularx}
\usepackage[export]{adjustbox}
\usepackage{amssymb}
\usepackage{pifont}
\usepackage{caption}
\usepackage{subcaption}
\usepackage{booktabs}
\usepackage{array}
\usepackage{xcolor}
\usepackage{xspace}
\usepackage{xspace}
\usepackage[normalem]{ulem}
\usepackage{algpascal}
\usepackage{inconsolata}
\usepackage{letltxmacro}
\usepackage{booktabs}
\usepackage{algorithm}
\usepackage{algpascal}
\usepackage{algpseudocode}
\usepackage{algorithmicx}
\usepackage{caption}
\usepackage{colortbl,xcolor,xspace}
\usepackage{soul}
\usepackage{nicematrix}
\usepackage{times}
\usepackage{latexsym}
\usepackage{graphicx}
\usepackage{multirow}
\usepackage{tabularx}
\usepackage[export]{adjustbox}
\usepackage{amssymb}
\usepackage{pifont}
\usepackage{longtable}
\usepackage{arydshln}
\usepackage{xtab,afterpage}
\usepackage[normalem]{ulem}

\definecolor{colone}{RGB}{133, 0, 3}
\newcommand{\colone}[1]{\textcolor{colone}{#1\xspace}}
\definecolor{coltwo}{RGB}{190, 81, 7}

\definecolor{colthree}{RGB}{13,85, 2}
\newcommand{\colthree}[1]{\textcolor{colthree}{#1\xspace}}
\definecolor{colfour}{RGB}{8, 0, 135}

\newcommand\blfootnote[1]{%
  \begingroup
  \renewcommand\thefootnote{}\footnote{#1}%
  \addtocounter{footnote}{-1}%
  \endgroup
}

\newcommand{\triple}[3]{\texttt{<#1, #2, #3>}}
\useunder{\uline}{\ul}{}

\title{ABEX: Data Augmentation for Low-Resource NLU via Expanding Abstract Descriptions}

\author{
    Sreyan Ghosh\textsuperscript{\rm*},
    Utkarsh Tyagi\textsuperscript{\rm*},
    Sonal Kumar\textsuperscript{\rm},
    Chandra Kiran Reddy Evuru\textsuperscript{\rm},\\
    \bf S Ramaneswaran\textsuperscript{\rm},
    \bf S Sakshi \textsuperscript{\rm},
    \bf Dinesh Manocha\textsuperscript{\rm} \\
    \textsuperscript{\rm}University of Maryland, College Park, USA \\
    \texttt{\{sreyang,utkarsht,sonalkum,ckevuru,ramans,fsakshi,dmanocha\}@umd.edu} \\
}

\begin{document}
\maketitle
\begin{abstract}
We present \textbf{ABEX}, a novel and effective generative data augmentation methodology for low-resource Natural Language Understanding (NLU) tasks. ABEX is based on \textbf{AB}stract-and-\textbf{EX}pand, a novel paradigm for generating diverse forms of an input document -- we first convert a document into its concise, abstract description and then generate new documents based on expanding the resultant abstraction. To learn the task of expanding abstract descriptions, we first train BART on a large-scale synthetic dataset with abstract-document pairs. Next, to generate abstract descriptions for a document, we propose a simple, controllable, and training-free method based on editing AMR graphs. ABEX brings the best of both worlds: by expanding from abstract representations, it preserves the original semantic properties of the documents, like style and meaning, thereby maintaining alignment with the original label and data distribution. At the same time, the fundamental process of elaborating on abstract descriptions facilitates diverse generations. We demonstrate the effectiveness of ABEX on 4 NLU tasks spanning 12 datasets and 4 low-resource settings. ABEX outperforms all our baselines qualitatively with improvements of 0.04\% - 38.8\%. Qualitatively, ABEX outperforms all prior methods from literature in terms of context and length diversity~\footnote{Code and data: https://github.com/Sreyan88/ABEX}.

\blfootnote{${^*}$Equal Technical Contribution.}
\end{abstract}

\section{Introduction}

Improving the performance of deep learning models on downstream Natural Language Understanding (NLU) tasks requires sufficient good-quality training data. However, data annotation is an expensive, time-consuming, and noisy task~\cite{abad-moschitti-2016-taking}. Data augmentation has proven to be an effective approach for overcoming the data scarcity issue in low-resource NLU tasks with limited training samples~\cite{chen2023empirical}. The two major categories of study in data augmentation include online data augmentation by interpolation in the latent space \cite{guo2019augmenting, ng2020ssmba, sun2020mixup,Kumar2020,guo2020nonlinear, sawhney2021hypmix} and offline data augmentation that expands an existing small-scale dataset by generating additional synthetic data \cite{wei2019eda, Kumar2020, zhou2021melm, kim2022alp, guo2022genius}. Owing to advancements in generative models that facilitate the creation of high-quality synthetic data, the latter is gaining traction~\cite{yu2023large}.

\begin{table}[t!]
    \centering
    \resizebox{1.0\columnwidth}{!}{%
    {\renewcommand{\arraystretch}{1.05}%
    \begin{tabular}{ >{\footnotesize} l | >{\small} l }
    \hline
         \multirow{3}{*}{\textbf{Method}}  & \textbf{Original 1:} \colthree{Usually, the two of us don't agree on anything about politics.}\\ 
         & \textbf{Original 2:} \colone{The pop superstar said she was "completely inspired" by} \\
         &   \colone{Roem's victory.} \\ \hline
        
        \multirow{2}{*}{EDA}       & 1. The two of us dont on about politics \\
        \multirow{2}{*}{(\citeauthor{wei2019eda})}& 2. \colone{Bulge} the pop superstar said she was completely inspired by roems  \\ 
        &  victory \\ \hline
        \multirow{2}{*}{AEDA}      & 1. Usually, the two of us \colthree{?} don't agree \colthree{;} on anything \colthree{!} about \colthree{:} politics.  \\
        \multirow{2}{*}{(\citeauthor{karimi-etal-2021-aeda-easier})}& 2. The pop superstar \colone{;} said she was ""completely inspired"" by Roem's \\ & victory.  \\ \hline
        \multirow{2}{*}{SSMBA}     & 1. Usually, the two of us don't agree about anything \colthree{involving} politics.   \\
        \multirow{2}{*}{(\citeauthor{ng2020ssmba})}& 2. The pop superstar said she \colone{felt} was completely inspired "" by roem\'s \\ & victory\colone{!} \\
        \hline
        \multirow{2}{*}{AMR-DA}     & 1. We usually don't agree on anything.   \\
        \multirow{2}{*}{(\citeauthor{shou-etal-2022-amr})}& 2.  Pop superstar\colone{s} \colone{say that a complete victory for} Roem \colone{and superstars} \\ & will inspire them . \\
        \hline
        \multirow{1}{*}{GENIUS}    & 1. It about politics. {It about everything.} \\
        \multirow{1}{*}{(\citeauthor{guo2022genius})}& 2. The pop superstar. \colone{The singer. The songwriter.} \\
        \hline
        \multirow{2}{*}{LLaMA-2\textsubscript{13B}} & 1. Political disagreement is \colthree{the norm} between the two of us.  \\
        \multirow{2}{*}{(\citeauthor{touvron2023LLaMA})}& 2. The pop star \colone{also noted} that Roem's triumph had inspired her own \\
        & \colone{creative process}. \\  \hline
        \multirow{2}{*}{ZeroGen}    & 1. The two of us may disagree on \colthree{anything}, but we do not agree on it. \\ 
        \multirow{2}{*}{(\citeauthor{ye-etal-2022-zerogen})} & point at hand. \\
        & 2. The pop \colone{icon expressed} being \colone{tremendously} inspired by Roem. \\
        \hline
        \multirow{5}{*}{\textbf{ABEX} \textit{(ours)}}      & 1. \colthree{President Obama has failed to reach an agreement on any political}     \\
        & \colthree{issues, including the Iran nuclear deal, and there is no consensus on} \\ & \colthree{the next steps.} \\
        & 2. \colone{Cristiano Ronaldo is} inspired by Roem's victory \colone{over Manchester} 
   \\ 
        & \colone{United, according to the Portuguese superstar.}\\
    \hline
    \end{tabular}
    }}
    \caption{\small Comparison of augmentations generated using ABEX and our baselines on a \textit{randomly chosen} document from HuffPost. (\textbf{1.} \colthree{Politics}, \textbf{2.} \colone{Entertainment}). ABEX moves beyond simple text-editing or rephrasing and generates diverse augmentations by introducing a new context. Augmentations by ABEX are also more coherent and label-consistent.}
    \label{tab:teaser-table}
\end{table}

However, generative data augmentation faces two major challenges: \textit{diversity} in generated augmentations~\cite{geiping2023how} and \textit{consistency} with the underlying data distribution \cite{chen2023empirical}. It is crucial to strike a balance between these two aspects, as overemphasizing one at the expense of the other can lead to poor downstream performance. Current augmentation methods based on text-infilling~\cite{ghosh-2023-aclm,guo2022genius,wang-etal-2022-promda}, where the primary task is to generate a new sentence constrained with keywords, are prone to replicate biases and overfit specific linguistic patterns in the low-resource training data, thereby hurting diversity. Additionally, we show that keyword-constrained free-form generation is unable to maintain the core semantic properties of the document, like style, which proves to be critical for specific tasks (e.g., \textit{question} style document for intent classification. See example in Table~\ref{fig:atis}). Diversity also proves to be an issue with token-level editing methods~\cite{wei2019eda,shou-etal-2022-amr} that rarely introduce novel entities or contexts and often randomly edits important tokens. Finally, prompt-based methods that employ Large Language Models (LLMs) require well-curated attributes selected from the data to control the distribution of the generated data~\cite{yoo-etal-2021-gpt3mix-leveraging,sahu2023promptmix,yu2023large}.
\vspace{0.5mm}

{\noindent \textbf{Main Contributions. }} In this paper, we propose \textbf{ABEX}, a novel data augmentation methodology based on a novel paradigm - Abstract-and-Expand. We first convert an input document into a concise, abstract description of itself and then generate augmentations by expanding the resultant abstraction. The task emulates human language perception and processing: the abstraction phase mirrors how humans distill core ideas from text, focusing on essential meanings, while the expansion phase reflects human creativity in generating varied narratives from a single abstract concept, akin to human extrapolation of ideas into diverse discussions. Our proposed Abstract-and-Expand task, which differs from all tasks proposed in prior art, generates augmentations that are both more consistent and diverse. To learn the task of expanding abstract descriptions, we first synthesize a large-scale synthetic dataset by prompting LLMs and then train an Encoder-Decoder Pre-trained Language Model (BART~\cite{lewis2019bart})  on the dataset. Next, we propose a simple and controllable algorithm to generate abstract descriptions for training instances in any given downstream low-resource dataset. Our proposed algorithm leverages AMR-to-Text and Text-to-AMR and generates abstract descriptions by editing Abstract Meaning Representation (AMR) graphs~\cite{banarescu2013abstract}. Inspired by the success of mixup in data augmentation~\cite{zhang2018mixup}, we also optionally mix AMR graphs of two sentences to boost the diversity of abstract descriptions. Finally, we synthesize diverse augmentations using the fine-tuned model and synthesized abstract descriptions. To summarize, our main contributions are:

\begin{enumerate}
    \item We propose ABEX, a novel and effective generative data augmentation methodology for low-resource NLP. We employ a novel Abstract-and-Expand task and fine-tune an Enc-Dec PLM to learn the task. ABEX differs from all prior work in its motivation and methodology and closely mimics the human perception and processing of language. 
    \item We propose a simple, controllable, and training-free method for generating abstract descriptions of source documents from downstream NLU datasets. Our proposed methodology provides explicit control in the document-to-abstract generation process and overcomes the contained generation issue that LLMs face in abstract generation.
    \item  To evaluate the efficacy of ABEX augmentations, we experiment on 12 datasets across 4 NLU tasks under 4 low-resource settings and show that ABEX outperforms most prior works quantitatively by 0.04\% - 38.8\%. Additionally, generations by ABEX are superior to prior work in terms of context, token (including entity), and length diversity.
    \item We also contribute the large-scale synthetic dataset with $\approx$0.2 million abstract-expansion pairs to promote further research in this space.
\end{enumerate}

\section{Background and Related Work}
\label{sec:related_background}

{\noindent \textbf{Definition of abstract description.}} An abstract description is a concise summary of a text, distilling it to its key concepts and themes while omitting non-essential details, effectively retaining the text's core message. Examples can be seen in Table~\ref{tab:examples_1}.
\vspace{0.5mm}

{\noindent \textbf{Difference between an abstract description and an (abstract) summary.}} A summary provides a concise overview of the main points or themes of a text, maintaining the original structure and order of ideas. In contrast, an abstract description distills the essence or core concept of the text, often rephrasing or reorganizing the content to capture its fundamental meaning in a more generalized form. In the case of summary generation, while including entities and primary events in the text is incentivized, abstract descriptions should only describe the broad semantic meaning of the text. Contrasting examples are in Tables~\ref{tab:examples_1} and \ref{tab:examples_2}.  
\vspace{0.5mm}

{\noindent \textbf{Background on AMR graphs.}} An AMR graph \cite{banarescu2013abstract} is a linguistic representation of a sentence that captures the meaning of a document in a structured manner. Formally put, an AMR graph can be represented as $\mathcal{G}$ = ($\mathcal{V}$, $\mathcal{E}$), where each vertex $\mathcal{V}$ represents a concept, and each edge $\mathcal{E}$ represents a relationship between concepts. 



\vspace{0.5mm}

{\noindent \textbf{Generative Data Augmentation for NLP.}}
Generative data augmentation for low-resource NLP can be broken down into 4 main categories: \textbf{(1)} Text-infilling: Given a source text, the task is to corrupt parts of the text and infill the corrupted parts using a Pre-trained Language Model (PLM). The task is generally completed by conditioning the corrupted text (also framed as keyword conditioning by some prior work) to an auto-regressive model~\cite{zhou2021melm,guo2022genius,ghosh-2023-aclm,ghosh-etal-2023-dale,10.1145/3539618.3591957}. The parts of the input text to be corrupted are either chosen randomly~\cite{Kumar2020} or algorithmically~\cite{guo2022genius,ghosh-2023-aclm}. \textbf{(2)} Text-editing: Given a source sentence, the task is to edit parts of the sentence~\cite{wei2019eda,shou-etal-2022-amr}. \textbf{(3)} Prompting: The task is to prompt LMs to generate novel training sentences~\cite{ye-etal-2022-zerogen,sahu2023promptmix}. The prompt may be further conditioned on attributes extracted from the training data, exemplars, or constraints extracted from the training data. \textbf{(4)} Style conversion: The task is to rephrase or change the style of the source sentence~\cite{chen-etal-2022-style,sharma2022systematic}. \citet{chen2023empirical} perform a large-scale evaluation comparing several augmentation methods.
\vspace{1mm}

\section{Methodology}
\label{subsec:method}
{\noindent \textbf{Overview.}} Fig.~\ref{fig:diagram} illustrates the entire workflow of generating augmentations with ABEX. The workflow has 2 major steps: \textbf{(1)} We first learn the task of expanding abstract descriptions by fine-tuning BART on a large-scale synthetic dataset. To accomplish this, we first synthesize a dataset $\mathcal{D}_{ab}$, with abstract-document pairs ($x^{ab}_i$,$y^{ab}_i$) by prompting LLMs on a large unlabeled dataset $\mathcal{D}_u$. \textbf{(2)} We then generate synthetic augmentations for a downstream NLU dataset $\mathcal{D}_{down}$ with document-label pairs ($x^{down}_i$,$y^{down}_i$) by first converting the documents into abstract descriptions and then employing the fine-tuned BART to generate multiple diverse expansions. Directly prompting LLMs for abstraction and expansion affects controllability, and we also show that it underperforms ABEX. 




\subsection{Learning to Expand Abstract Descriptions}
\label{subsec:learning}

\begin{figure*}[t]
\centering
\includegraphics[width=2.0\columnwidth]{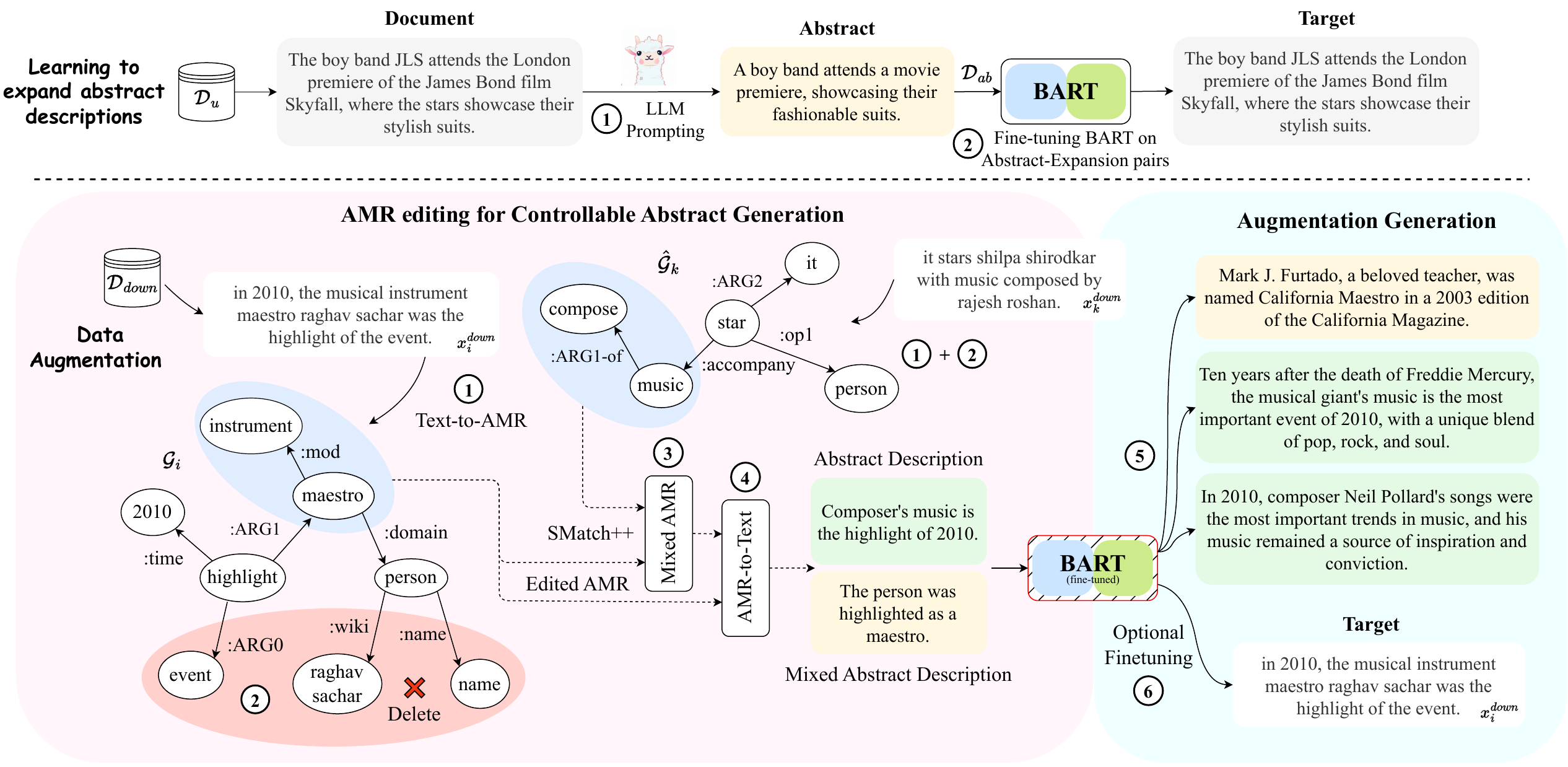}
\caption{\small Illustration of our proposed augmentation methodology. \textbf{Top: Learning to Expand Abstract Descriptions.} \textcircled{\raisebox{-0.9pt}{1}} We synthesize a large-scale synthetic dataset $\mathcal{D}_{ab}$ with abstract-document pairs by prompting LLMs with unlabeled documents from $\mathcal{D}_{ab}$. \textcircled{\raisebox{-0.9pt}{2}} We pre-train BART on this dataset with abstract as input and document as the target for learning to expand abstract descriptions. \textbf{Bottom: Data Augmentation.} \textcircled{\raisebox{-0.9pt}{1}} We convert the document into its AMR graph representation $\mathcal{G}_{i}$ using a Text-to-AMR Parser. \textcircled{\raisebox{-0.9pt}{2}} $\mathcal{G}_{i}$ then goes through multiple steps of \textit{deletion} to obtain $\hat{\mathcal{G}}_{i}$ \textcircled{\raisebox{-0.9pt}{3}} We optionally retrieve a semantically similar document from $\mathcal{D}_{down}$, obtain its AMR graph $\mathcal{G}_{k}$, and replace subtrees in $\hat{\mathcal{G}}_{i}$ with \textit{similar} subtrees in $\hat{\mathcal{G}}_{i}$. \textcircled{\raisebox{-0.9pt}{4}} $\hat{\mathcal{G}}_{i}$ is then converted back to text (which is now an abstract description) using an AMR-to-Text generator. \textcircled{\raisebox{-0.9pt}{5}} This abstract description is then passed to the fine-tuned BART for generating augmentations. \textcircled{\raisebox{-0.9pt}{6}} We optionally fine-tune the fine-tuned BART (from the 1st step) on abstract-document pairs from $\mathcal{D}_{down}$.}
\label{fig:diagram}
\end{figure*}

In this subsection, we provide an overview of the upper half in Fig.~\ref{fig:diagram}. We describe how we synthesize the synthetic dataset $\mathcal{D}_{ab}$ and fine-tune BART on this dataset to obtain a model capable of expanding abstract descriptions.
\vspace{0.5mm}

{\noindent \textbf{(1) Generating a synthetic dataset ($\mathcal{D}_{ab}$).}} Due to the lack of open-source datasets available for the task, we generate high-quality synthetic data for learning this task by prompting LLMs. We prompt an LLM with documents from $\mathcal{D}_u$ and ask it to generate an abstract description of them. However, the primary challenge in the proposed generation process is the choice of seed unlabeled datasets. Large-scale open-source datasets consist of long documents, in contrast to the nature of instances in the majority of downstream fine-tuning datasets that are made of much shorter documents. Mismatch in the length of training and inference datasets have been shown to degrade performance in various tasks in prior art~\cite{rogers2021primer,ghosh-etal-2023-dale}. The other alternative is to select individual sentences from these long documents. However, this creates an informativeness mismatch as individual and context-less sentences from these documents are rarely self-contained, unlike sentences in downstream datasets. Thus, to overcome these issues, we follow a two-step prompting strategy: \textbf{(i)} We first generate summaries of the original long documents in $\mathcal{D}_{u}$ \textbf{(ii)} We then generate abstract descriptions of each summary. We denote our final synthetic dataset by $\mathcal{D}_{ab}$, and $\mathcal{D}_{ab}$ is made of abstract-document pairs ($a$,$d$) where $a$ is the final output of the LLM from step \textbf{(ii)} and $d$ is the output from step \textbf{(i)}. An example can be seen in Fig.~\ref{fig:diagram}, and more examples are available in Tables~\ref{tab:examples_1} and \ref{tab:examples_2}. We employ LLaMA-2 13B~\cite{touvron2023LLaMA} for this task and generate $\approx$0.2 million abstract-document pairs for fine-tuning. Prompts are listed in Appendix~\ref{sec:prompt}. 
\vspace{0.5mm}

{\noindent \textbf{(2) Fine-tuning BART on $\mathcal{D}_{ab}$.}} After generating paired data, we fine-tune BART on $\mathcal{D}_{ab}$ to learn the task of expanding abstract descriptions. The abstract $a$ and the document $d$ serve as the input and target, respectively. 



\vspace{0.5mm}

\subsection{Data Augmentation using ABEX}
\label{subsec:augmenting_downstream}

This section provides an overview of the lower half in Fig.~\ref{fig:diagram}. The primary aim is to generate multiple diverse augmentations of every source document in the downstream task dataset $\mathcal{D}_{down}$, which can then be added to $\mathcal{D}_{down}$ to improve downstream task performance. We first generate abstract descriptions for each instance in $\mathcal{D}_{down}$ in a controlled manner using our proposed method (described next), followed by employing fine-tuned BART from step (1) to generate multiple expansions of the abstractions. These expansions then act as augmentations. 


\subsubsection{Controllable Generation of Abstract descriptions for $\mathcal{D}_{down}$}
\label{subsec:sentence_to_abstract}



{\noindent \textbf{Primary Motivation.}} The most straightforward method to generate abstract descriptions for each instance $x^{down}_i$ in $\mathcal{D}_{down}$ would have been to employ an LLM with the same prompt discussed in Section~\ref{subsec:learning}. However, there are 2 major challenges with this approach: 
\vspace{0.5mm}

{\noindent \textbf{(1) Maintaining Label Consistency.}} A key requirement of effective augmentations is that they maintain label consistency with the underlying Gold-only  training instance. For example, a synthetic augmentation of an instance from a sequence classification dataset with a label: \textit{positive sentiment} should also be of \textit{positive sentiment}. Prior data augmentation methods based on text-infilling usually retain target-related information (TRI) (or phrases relevant to the label) in the corrupted sentence, followed by infilling text around the TRI to generate augmentations~\cite{guo2022genius,ghosh-etal-2023-dale,ghosh-2023-aclm}. Inspired by this, our primary motive is to generate an abstract description of $x^{down}_i$ that retains the TRI corresponding to its label $y^{down}_i$. Doing this would also ensure that the expansion (or augmentations) would be label-consistent. Accomplishing this using the prompting method discussed in Section~\ref{subsec:learning} would require the LLM to be effective at constrained generation. Recent studies, such as the work by ~\citet{lu2023bounding} and ~\citet{sun-etal-2023-evaluating}, suggest that while constrained generation can make prompts more complex, it may also present challenges for LLMs in consistently adhering to the constraints mentioned in prompts.
\vspace{0.5mm}

{\noindent \textbf{(2) Controlling the degree of abstraction.}} The degree of abstraction for generating abstract descriptions affects the final augmentations in terms of diversity and label consistency. These factors, in turn, affect downstream performance, and the optimal degree of abstraction varies from task to task. Similar to the above, controlling the degree of abstractions proves to be difficult for LLMs. Additionally, the nature of TRIs differs from task to task, which increases the complexity of the prompts significantly.

{\noindent \textbf{Proposed Solution.}} To overcome the controlled generation bottleneck in LLMs, we propose a simple yet controllable and effective method for generating abstract descriptions. Based on AMR editing, our proposed method is \textit{training-free} and essentially performs text-editing, so there is no need to learn a model for every dataset. Additionally, it is flexible and can easily cater to a wide range of tasks without significant algorithmic changes.






{\noindent \textbf{(1) Text - to - AMR.}} Our first step is to convert a document into its AMR graph. To perform this step, we employ text-to-AMR AMR-BART \cite{bai-etal-2022-graph}, which is built on BART and trained to generate AMR graphs from text.
\vspace{0.5mm}

{\noindent \textbf{(2) Editing the AMR.}} Following the definition of abstract descriptions and AMRs in Section~\ref{sec:related_background}, editing AMR graphs provides a feasible way to generate an abstract description by deleting nodes corresponding to specific, non-central details and keeping the ones that capture the meaning and essence. The editing operations are designed such that the edited AMR graph, once converted back to text, results in an abstract description of the original document. We first linearize the AMR graph generated in Step 1 into a sequence \cite{bai-etal-2022-graph} to achieve this. However, before editing, we want to ensure we retain the original TRI for the document in the AMR. Thus, inspired by \citet{ghosh-etal-2023-dale} and \citet{guo2022genius}, we first extract top-\textit{k} keywords in the document by measuring the similarity between n-grams from the document and the document label. Once extracted, we ensure these keywords are not edited in the AMR. Note that TRI extraction differs from task to task, and we request that our readers refer to our code for more details.


Next, we perform multiple rounds of \textit{deletion} operation on the AMR graph. First, we remove certain pre-defined types of attributes from the AMR. Some examples of these types are $:value$, $:wiki$, $:mod$ and $:quant$. We list all such attributes that serve as our candidates for the deletion operation in Appendix \ref{subsec:amr_attributes}. After attribute deletion, we then delete sub-graphs in the AMR graph. A sub-graph can be seen as a broader conceptual unit describing a specific idea entailed to a concept or entity. Deleting a sub-graph leads to a higher level of abstraction, thereby leading to more diverse sentences (ablation in \ref{sec:p_gen}). We select our candidate subgraphs for deletion based on a metric we define as the \textit{depth-ratio}. To calculate the depth ratio, we calculate the ratio of the depth of the sub-graph to the entire graph. We define \textit{depth} as measuring the distance between the root node and the farthest leaf node. Specifically, it captures the vertical span and the nesting level within an AMR graph. We select a sub-graph as an eligible candidate for deletion only if its depth ratio is less than a given threshold $\alpha$. The maintenance of a depth ratio enables us to regulate the size of the removed graph, thereby determining the level of abstraction. We then sample a deletion rate $\varepsilon$ from a Gaussian distribution $\mathcal{N}(\mu,\,\sigma^{2})$ and dynamically delete $\varepsilon\%$ sub-graphs among eligible candidates.
\vspace{0.5mm}



{\noindent \textbf{(3) Mixing AMR graphs of 2 documents.}} Mixing samples in the training data to generate new data with concepts from both samples has been a successful augmentation approach across modalities~\cite{zhang2018mixup,sahu2023promptmix}. The method, also commonly known as \textit{mixup}, improves the diversity of generated data through semantic interpolation, which in turn leads to more generalized models. To perform mixup in the ABEX framework, we can generate abstract descriptions with mixed concepts from a pair of training instances and then employ $\mathcal{B}$ for diverse expansions. Formally, let $x^{down}_i$ be the source document and $x^{down}_k$ be another retrieved sentence that is semantically similar to $i_{n}$. We retrieve $x^{down}_k$ using cosine similarity with SentenceBERT~\cite{reimers2019sentence}. After editing the AMR graphs, $\mathcal{G}_{i}$ and $\mathcal{G}_{k}$, of documents $x^{down}_i$ and $x^{down}_k$ respectively, we first extract all their possible sub-graphs from both AMR graphs. Each sub-graph intuitively represents an individual concept in an AMR graph. We denote the set of sub-graphs as $\mathcal{S}^{i}$ and $\mathcal{S}^{k}$, where $\mathcal{S}^{i} = \{s^{i}_0, \cdots, s^{i}_n\}$ and $n$ is the total number of sub-graphs (similar for $\mathcal{S}^{k}$). We now calculate the sub-graph similarity between each pair of sub-graphs in $\mathcal{S}^{i}$ and $\mathcal{S}^{k}$ and append the top-$k$ sub-graphs in $\mathcal{S}^{k}$ to their most similar to sub-graphs $\mathcal{S}^{i}$. To calculate sub-graph similarity, we employ SMATCH++ \cite{opitz-2023-smatch} at the sub-graph level (details on SMATCH++ in Appendix \ref{subsec:similar_retreival}). The resultant AMR graph $\hat{\mathcal{G}_{i_{n}}}$ is then used in Step 4. For generating $R \times$ augmentations of $x^{down}_i$, we do not apply this step on all rounds $R$ but sample a probability $\gamma$ from a Gaussian distribution $\mathcal{N}(\mu,\,\sigma^{2})$ and only apply this if $\gamma$ crosses a set threshold $\beta$. 
\vspace{0.5mm}

{\noindent \textbf{(4) AMR - to - Text.}} To convert the edited graph back to text, we employ AMR-to-text AMR-BART~\cite{bai-etal-2022-graph}. For our experiments, we employ pre-trained checkpoints provided by the authors in their code release.

\subsubsection{Augmentation Generation}

{\noindent \textbf{Optional Fine-tuning on $\mathcal{D}_{down}$.}} We optionally fine-tune the fine-tuned BART (from the 1st step) on the low-resource downstream dataset for domain adaptation.  To obtain abstract-document pairs for this step, we employ the methodology defined in Section~\ref{subsec:sentence_to_abstract} to generate abstracts for each document in the downstream dataset but skip Step (3) (note that mixing AMR graphs of 2 sentences in  Step (3) voids the relationship of the abstract with the original document). 
\vspace{0.5mm}

{\noindent \textbf{Generation.}} After optional fine-tuning, we feed the generated abstracts from $\mathcal{D}_{down}$ to the fine-tuned BART capable of expanding abstract descriptions and generating diverse expansions that serve as augmentations. To boost diversity, during auto-regressive generation, we perform random multinomial sampling and sample the next word from the top-\textit{k} most probable words and choose the most probable sequence with beam search. For generating $R\times$ synthetic data, we repeat this process for $R$ rounds and add the synthetic augmentations with the Gold-only data for training the downstream NLU model. Note that post fine-tuning BART on $\mathcal{D}_{ab}$, ABEX can be considered as a training-free data augmentation method, i.e., ABEX does not require fine-tuning for specific downstream datasets. Fine-tuning on $\mathcal{D}_{down}$ is optional, and generating abstracts only requires pre-trained models.

\begin{table*}[]
\small
\centering
{\renewcommand{\arraystretch}{1.15}%
\resizebox{\linewidth}{!}{%
\begin{tabular}{l|cccc|cccc|cccc|cccc|cccc}
\toprule \toprule
\multicolumn{1}{c|}{\multirow{2}{*}{Model}}                     & \multicolumn{4}{c|}{Huffpost}                                                                                                 & \multicolumn{4}{c|}{Yahoo}                                                                                                     & \multicolumn{4}{c|}{IMDB}                                                                                                      & \multicolumn{4}{c|}{\textbf{ATIS}}                                                                                                      & \multicolumn{4}{c}{MASSIVE}                                                                                                    \\
\multicolumn{1}{c|}{}                                           & \textbf{100} & \textbf{200} & \textbf{500} & \textbf{1000} & \textbf{100} & \textbf{200} & \textbf{500} & \textbf{1000} & \textbf{100} & \textbf{200} & \textbf{500} & \textbf{1000} & \textbf{100} & \textbf{200} & \textbf{500} & \textbf{1000} & \textbf{100} & \textbf{200} & \textbf{500} & \textbf{1000} \\ \midrule
Gold                                                            & 76.80                          & 77.96                         & 80.51                         & 82.41                          & 42.50                         & 49.50                         & \cellcolor{magenta!20}\textbf{55.47}                & 56.62                          & 83.36                         & {\ul 88.59}                   & 88.15                         & \cellcolor{magenta!20}\textbf{89.47}                 & 85.13                         & 89.97                         & 94.7                          & 97.29                          & 31.70                         & 56.48                         & 73.47                         & 79.15                          \\
BackTrans                                                       & 75.87                         & 76.21                         & 79.20                         & 80.20                          & 44.85                         & 50.86                         & 54.19                         & 55.77                          & {\ul 84.38}                   & 86.12                         & 86.72                         & 87.53                          & 89.86                         & 92.34                         & 94.36                         & 97.07                          & {\ul 53.56}                   & 64.52                         & 73.13                         & 78.48                          \\
EDA                                                             & 75.49                         & 77.64                         & 79.14                         & 80.71                          & 47.13                         & 50.15                         & 53.39                         & 56.04                          & 75.3                          & 88.07                         & 88.39                         & 88.92                          & 90.20                         & 92.11                         & 94.93                         & 96.62                          & 47.00                         & 64.15                         & 73.53                         & 78.24                          \\
AEDA                                                            & 77.65                         & 76.88                         & 80.31                         & 81.10                          & 45.61                         & 51.52                         & 54.22                         & 56.02                          & 82.30                         & 88.25                         & 86.95                         & {\ul 89.33}                    & 89.07                         & 91.89                         & 96.73                         & {\ul 97.63}                    & 51.04                         & {\ul 66.81}                   & {\ul 75.15}                   & 79.11                          \\
AMR-DA                                                          & 77.49                         & 76.32                         & 77.93                         & 79.64                          & 48.80                         & 52.37                         & 54.68                         & 55.01                          & 84.26                         & 88.04                         & {\ul 88.92}                   & 89.20                          & {\ul 93.69}                   & 94.03                         & 96.28                         & 96.39                          & 52.82                         & 64.02                         & 72.09                         & 76.96                          \\
SSMBA                                                           & 76.64                         & 77.40                         & 79.85                         & 81.11                          & 46.95                         & 50.53                         & 53.97                         & 54.68                          & 82.09                         & 86.57                         & 87.94                         & 88.8                           & 90.31                         & 89.75                         & 93.69                         & 95.94                          & 47.07                         & 60.99                         & 70.24                         & 77.16                          \\
GENIUS                                                          & 77.52                         & 77.71                         & 78.35                         & 80.07                          & 51.9                          & 51.69                         & 51.46                         & 54.15                          & 78.58                         & 82.50                         & 84.90                         & 86.18                          & 93.58                         & 94.14                         & 96.73                         & 97.18                          & 51.76                         & 65.34                         & 73.17                         & 77.04                          \\
PromDA                                                          & {\ul 77.83}                   & 77.90                         & 77.65                         & 81.06                          & {\ul 52.61}                   & 52.13                         & 53.40                         & 56.27                          & 84.21                         & 88.24                         & 88.30                         & 88.65                          & -                             & -                             & -                             & -                              & -                             & -                             & -                             & -                              \\
PromptMix                                                       & -                             & -                             & -                             & -                              & -                             & -                             & -                             & -                              & -                             & -                             & -                             & -                              & 92.68                         & 94.25                         & 94.81                         & 96.95                          & 52.60                         & 64.53                         & 74.26                         & 76.87                          \\
ZeroGen                                                         & 73.84                         & 75.66                         & 76.30                         & 76.49                          & 41.47                         & 49.21                         & 54.55                         & 55.04                          & 76.99                         & 80.61                         & 82.31                         & 83.10                          & 81.24                         & 83.95                         & 85.63                         & 90.88                          & 28.20                         & 47.02                         & 67.80                         & 70.94                          \\
LLaMA-2\textsubscript{13B} & 73.59 & 75.19 & 76.82 & 77.94 & 40.37 & 46.25 & 52.14 & 53.62 & 80.72 & 83.59 & 85.62 & 85.81 & 82.80 & 81.72 & 89.11 & 91.05 & 30.88 & 49.19 & 70.52 & 71.80 \\
GPT3Mix                                                         & 57.87                         & 61.80                         & 66.12                         & 69.46                          & 31.60                         & 32.98                         & 50.33                         & 52.93                          & 81.04                         & 84.14                         & 86.27                         & 87.69                          & 76.91                         & 81.75                         & 85.36                         & 85.36                          & 25.91                         & 46.72                         & 68.99                         & 72.57                          \\
\hdashline
\textbf{ABEX-Abs} & 73.62 & 74.58 & 76.27 & 78.42 & 35.87 & 37.93 & 48.47 & 50.36 & 74.69 & 80.28 & 82.66 & 82.51 & 78.53 & 80.27 & 83.54 & 86.49 & 30.71 & 51.62 & 68.88 & 75.26  \\ 

\textbf{ABEX-stage-2}      & 74.61                         & 77.26                         & 78.17                         & 80.28                          & 49.81                         & 50.02                         & 51.62                         & 53.74                          & 82.69                         & 85.36                         & 87.22                         & 87.45                          & 90.71                         & 92.36                         & 96.75                         & 96.68                          & 50.47                         & 65.38                         & 73.29                         & 76.25                          \\
\textbf{ABEX-stage-1}                               & 77.45                         & {\ul 79.24}                   & {\ul 81.63}                   & {\ul 83.58}                    & 52.46                         & {\ul 53.26}                   & 54.77                         & \cellcolor{magenta!20}\textbf{57.13}                 & 84.35                         & 88.16                         & 88.30                         & 89.17                          & 91.66                         & {\ul 94.83}                   & {\ul 96.79}                   & 96.45                          & 52.51                         & 65.63                         & 73.94                         & {\ul 79.41}                    \\
\textbf{ABEX} \textit{(ours)} & \cellcolor{magenta!20}\textbf{78.66}                & \cellcolor{magenta!20}\textbf{79.30}                & \cellcolor{magenta!20}\textbf{81.82}               & \cellcolor{magenta!20}\textbf{84.03}                 & \cellcolor{magenta!20}\textbf{53.20}                & \cellcolor{magenta!20}\textbf{53.52}                & {\ul 54.81}                   & {\ul 57.11}                    & \cellcolor{magenta!20}\textbf{85.18}                & \cellcolor{magenta!20}\textbf{88.72}                & \cellcolor{magenta!20}\textbf{89.05}                & 89.28                          & \cellcolor{magenta!20}\textbf{94.28}                & \cellcolor{magenta!20}\textbf{95.71}                & \cellcolor{magenta!20}\textbf{97.33}                & \cellcolor{magenta!20}\textbf{97.92}                 & \cellcolor{magenta!20}\textbf{55.03}                & \cellcolor{magenta!20}\textbf{66.85}                & \cellcolor{magenta!20}\textbf{75.44}                & \cellcolor{magenta!20}\textbf{80.36}                 \\ 
 & $\pm0.72$               & $\pm0.05$               & $\pm0.13$              & $\pm0.42$                 & $\pm0.56$                & $\pm0.24$                & $\pm0.51$                   & $\pm0.01$                    & $\pm0.73$                & $\pm0.12$               & $\pm0.10$                & $\pm0.12$                          & $\pm0.54$                & $\pm0.78$                & $\pm0.45$                & $\pm0.24$                 & $\pm1.34$                & $\pm0.02$                & $\pm0.24$                & $\pm0.85$                 \\ 
 \bottomrule           
\end{tabular}
}}
\caption{\small Result comparison on Sequence Classification. ABEX outperforms prior methods by 0.04\% - 29.12\%.}
\label{tab:classification}
\end{table*}
\begin{table*}[t]
\begin{minipage}{0.5\textwidth}
\scriptsize
\centering
\resizebox{\textwidth}{!}{
\setlength{\tabcolsep}{1.8pt}
{\renewcommand{\arraystretch}{1.068}%
\begin{tabular}{l|cccc|cccc}
\toprule \toprule
\multicolumn{1}{c|}{\multirow{2}{*}{Model}}                     & \multicolumn{4}{c|}{MRPC}                                                                                                & \multicolumn{4}{c}{QQP} \\ 
\multicolumn{1}{c|}{}                                           & \textbf{100} & \textbf{200} & \textbf{500} & \textbf{1000} & \textbf{100} & \textbf{200} & \textbf{500} & \textbf{1000} \\ \midrule
                     Gold-only                                                                            & 66.47                         & 73.25                         & 77.55                         & 77.49                          & 69.23                         & 72.00                         & 75.27                         & 76.15                          \\
BackTrans                                                                      & 64.86                         & 71.01                         & 69.85                         & 69.68                          & 67.21                         & 69.44                         & 71.43                         & 72.34                          \\
EDA                                                                            & 65.56                         & 72.28                         & 74.55                         & 76.23                          & 69.22                         & 69.51                         & 70.64                         & 73.02                          \\
AEDA                                                                           & 62.43                         & 71.59                         & 74.84                         & 77.44                          & 69.45                         & 68.81                         & 72.54                         & 76.32                          \\
SSMBA                                                                          & 64.96                         & 70.82                         & 73.60                         & 75.23                          & 66.51                         & 63.10                         & 69.60                         & 70.73                          \\
AMR-DA                                                                         & 65.78                         & 73.10                         & 75.62                         & 77.02                          & 69.58                         & 70.63                         & 72.31                         & 73.66                          \\
LLaMA-2\textsubscript{13B} & 66.21 & 72.55 & 76.72 & 77.78 & 70.35 & 73.57 & 74.39 & 74.81 \\
\hdashline                    
\textbf{ABEX-Abs} & 63.52 & 70.71 & 75.46 & 76.21 & 68.31 & 70.44 & 72.30 & 73.08  \\
\textbf{ABEX-stage-2} & 66.59                         & 73.88                         & 77.24                         & 77.58                          & 70.24                         & 71.68                         & 74.57                         & 74.89                          \\
\textbf{ABEX-stage-1}             & {\ul 68.17}                   & \cellcolor{magenta!20}\textbf{74.36}                & {\ul 77.92}                   & {\ul 78.04}                    & {\ul 71.60}                   & {\ul 74.02}                   & {\ul 76.49}                   & {\ul 76.73}                    \\
\textbf{ABEX} \textit{(ours)}                & \cellcolor{magenta!20}\textbf{68.36}                & {\ul 74.29}                   & \cellcolor{magenta!20}\textbf{78.11}                & \cellcolor{magenta!20}\textbf{78.36}                 & \cellcolor{magenta!20}\textbf{72.13}                & \cellcolor{magenta!20}\textbf{74.32}                & \cellcolor{magenta!20}\textbf{76.53}                & \cellcolor{magenta!20}\textbf{76.81}                
 \\
                    & $\pm0.37$ & $\pm0.32$ & $\pm0.73$ & $\pm 0.21$ & $\pm0.55$ & $\pm0.28$ & $\pm0.86$ & $\pm0.62$
                                \\\bottomrule 
\end{tabular}%
}
}
\caption{\small Result comparison on Sentence Similarity. ABEX outperforms our baselines by 0.48\% - 11.22\%.}
\label{tab:sentence_sim}
\end{minipage}\hspace{0.02\textwidth}
\begin{minipage}{0.5\textwidth}
\scriptsize
\centering
\resizebox{\textwidth}{!}{
\setlength{\tabcolsep}{1.8pt}
{\renewcommand{\arraystretch}{1.068}%
    \begin{tabular}{l|cccc|cccc}
\toprule \toprule
\multicolumn{1}{c|}{\multirow{2}{*}{Model}}                     & \multicolumn{4}{c|}{SQuAD}                                                                                                & \multicolumn{4}{c}{NewsQA} \\ 
\multicolumn{1}{c|}{}                                           & \textbf{100} & \textbf{200} & \textbf{500} & \textbf{1000} & \textbf{100} & \textbf{200} & \textbf{500} & \textbf{1000} \\ \midrule
                     Gold-only  & 11.64                         & 19.71                         & 26.32                         & 31.52  &  22.45             & 30.14             & 45.65             & 58.83   \\
                     BackTrans  & 17.47                     & 22.60                         & 29.07                         & 32.60  & 27.32             & 34.98             & 47.21             & 60.21 \\
                     EDA      & 17.07                      & 22.39                         & 28.98                         & 32.40 & 29.31              & 35.81             & 49.90             & 61.01\\
                     AEDA       & 17.95                         & 23.50                        & 29.20                         & 32.68    & 29.87              & 36.80             & 50.24             & 61.78 \\
                     SSMBA      & 16.97                      & 22.27                         & 28.51                         & 32.01 & 28.89             & 33.27             &  47.56            & 60.34\\ 
        GENIUS      & 33.15                         & 42.65                         & 56.52                         & 65.62   & 38.88             &  47.36            & 57.32             & 69.36 \\
LLaMA-2\textsubscript{13B} & 34.62 & 42.58 & 58.92 & 65.71 & 40.86 & 50.24 & 56.58 & 68.97 \\
\hdashline
\textbf{ABEX-Abs} & 22.16 & 25.77 & 31.85 & 42.63 & 32.09 & 38.71 & 46.29 & 60.11 \\
\textbf{ABEX-stage-2}    & 35.67             & 45.34             &  58.79            & 66.23     & 41.78             & 49.82             & 57.38             & 71.63 \\
\textbf{ABEX-stage-1}                  & {\ul37.92}             & {\ul48.32}             & {\ul61.02}   & {\ul67.99}     & {\ul43.65}             & {\ul52.83}             & {\ul59.28}             & {\ul72.45} \\ 
\textbf{ABEX} \textit{(ours)}    &  \cellcolor{magenta!20}\textbf{38.34}            & \cellcolor{magenta!20}\textbf{49.87}             &  \cellcolor{magenta!20}\textbf{63.46}            &  \cellcolor{magenta!20}\textbf{70.32}           & \cellcolor{magenta!20}\textbf{45.75}             & \cellcolor{magenta!20}\textbf{54.67}             & \cellcolor{magenta!20}\textbf{61.43}             & \cellcolor{magenta!20}\textbf{73.41} \\
& $\pm0.21$ & $\pm0.19$ & $\pm0.70$ & $\pm 0.34$ & $\pm0.44$ & $\pm0.18$ & $\pm0.56$ & $\pm0.42$ \\
\bottomrule 
\end{tabular}%
    }
}
\caption{\small Result comparison on QA. ABEX outperforms all our baselines by 4.05\% - 38.8\%.}
\label{tab:qa}
\end{minipage}
\end{table*}

\section{Experimental Setup}
\label{subsec:exp}

\subsection{Tasks and Datasets}
\label{subsec:task_dataset}

{\noindent \textbf{Upstream Fine-tuning Dataset.}} For learning to expand abstract descriptions, we employ $\mathcal{D}_{ab}$ which consists of ~0.2 million unique abstract-document pairs.
\vspace{0.5mm}

{\noindent \textbf{Downstream Fine-tuning Datasets.}} To evaluate the efficacy of ABEX augmentations on downstream low-resource NLU tasks, we are largely inspired by the evaluation setup followed by a wealth of prior work in data augmentation~\cite{sahu2023promptmix,wang-etal-2022-promda,guo2022genius,ye-etal-2022-zerogen}. We additionally evaluate ABEX on the NER task, which prior work does not. Specifically, we evaluate 12 challenging datasets across 4 NLU tasks under 4 low-resource settings as follows:

For \textit{Sequence Classification} (SC) task, we employ Huffpost~\cite{huffpost} (news category classification), IMDB~\cite{maas-EtAl:2011:ACL-HLT2011} and Yahoo!\cite{zhang2015character} (answer topic classification), and ATIS~\cite{coucke2018snips} and Massive~\cite{fitzgerald2022massive} (intent classification). 

For \textit{NER}, we employ ConLL-2003~\cite{tjong-kim-sang-de-meulder-2003-introduction},  OntoNotes-5.0~\cite{pradhan2013towards} and MultiCoNER \cite{malmasi-etal-2022-multiconer} datasets, where all have a common set of tags and some unique tags. 

For the Question Answering (QA), we employ SQuAD~\cite{rajpurkar2016squad} and NewsQA~\cite{trischler-etal-2017-newsqa}. 

For the \textit{Sentence Similarity} (SS), we employ MRPC \cite{dolan2005automatically} and the Quora Question Pairs (QQP) dataset. 

Finally, to show that ABEX does not replicate spurious correlations from the training data in the generated augmentations, we employ SNLI~\cite{bowman-etal-2015-large} and MNLI~\cite{williams-etal-2018-broad}. These two datasets are known to have spurious correlations. We evaluate on the hard subsets of the test set in a setting similar to \citet{wu-etal-2022-generating}. Appendix \ref{sec:dataset_details} provides more details and statistics about these datasets.

\subsection{Hyper-parameters}
\label{subsec:hyper}

We employ BART\textsubscript{large} for learning to expand abstract descriptions. Our choice is motivated by the popularity of BART\textsubscript{large} in data augmentation literature~\cite{ghosh-etal-2023-dale,ghosh-2023-aclm,wang-etal-2022-promda}. We train it 15 epochs using Adam optimizer with a fixed learning rate of $5.6e^{-5}$.
For downstream NLU fine-tuning, we employ BERT\textsubscript{base-cased}~\cite{chalkidis-garneau-etal-2023-lexlms}. We fine-tune for 100 epochs with a batch size of 4,8 for 100 and 200 splits and 16 for 500 and 1000 splits. For SC and QA, we use Adam optimizer with a fixed learning rate of $1e^{-5}$. For NER, we employ FLAIR~\cite{akbik2019flair} with a starting lr of $1e^{-5}$ and constant decay. For AMR editing, we set $\mu$, $\sigma^2$,  and $\alpha$ to be 0.5, 0.1, and 0.35, respectively. For AMR mixing, we set $\mu$, $\sigma^2$,  and $\beta$ to be 0.5, 0.1, and 0.6, respectively. Appendix~\ref{sec:hyper-parameter} provides hyper-parameter tuning experiments. For low-resource experiments, we perform iterative stratified sampling over the dataset across four low-resource settings: 100, 200, 500, and 1000. We generate $R$=5 augmentations for all baselines and ABEX for all our experiments. We downsample the development set accordingly. We report the micro-average F$_1$ score averaged across 3 runs for 3 random seeds. We provide model results on hyper-parameter tuning in Appendix~\ref{sec:hyper-parameter}.

\subsection{Baselines}
\label{subsec:baselines}

\begin{table*}[]
\small
\centering
{\renewcommand{\arraystretch}{1.05}%
\resizebox{\linewidth}{!}{%
\begin{tabular}{l|cccc|cccc|cccc}
\toprule \toprule
\multicolumn{1}{c|}{\multirow{2}{*}{Model}}                     & \multicolumn{4}{c|}{CoNLL-2003}                                                                                                & \multicolumn{4}{c|}{MultiCoNER}                                                                                                & \multicolumn{4}{c}{OntoNotes}                                                                                                  \\
\multicolumn{1}{c|}{}                                           & \textbf{100} & \textbf{200} & \textbf{500} & \textbf{1000} & \textbf{100} & \textbf{200} & \textbf{500} & \textbf{1000} & \textbf{100} & \textbf{200} & \textbf{500} & \textbf{1000} \\ \midrule
Gold-only                                                            & 52.89                         & 66.53                         & 70.43                         & 80.15                          & 15.86                         & 24.91                         & 52.69                         & {\ul 57.03}                    & 16.37                         & 27.7                          & {\ul 61.46}                   & 61.82                          \\
LwTR                                                            & 65.48                         & {\ul 73.24}                   & {\ul 81.45}                   & {\ul 83.74}                    & 42.23                         & {\ul 50.22}                   & 51.0                          & 54.67                          & 46.18                         & \cellcolor{magenta!20}\textbf{51.47}                & 54.87                         & {\ul 62.67}                    \\
DAGA                                                            & 53.91                         & 51.63                         & 54.68                         & 82.05                          & 19.11                         & 36.71                         & 31.39                         & 42.13                          & 33.29                         & 43.07                         & 54.64                         & 61.15                          \\
MELM                                                            & 56.89                         & 62.23                         & 79.05                         & 81.90                          & 16.62                         & 30.96                         & 46.27                         & 49.01                          & 11.94                         & 31.55                         & 45.68                         & 54.97                          \\
GENIUS                                                          & 67.85                         & 58.2                          & 80.36                         & 76.87                          & {\ul 42.33}                   & 47.77                         & {\ul 55.70}                   & 51.06                          & 45.44                         & 48.69                         & 52.27                         & 56.59                          \\
PromDA                                                          & 66.30                         & 70.95                         & 76.38                         & 82.14                          & 41.40                         & 48.93                         & 55.02                         & 53.55                          & 46.34                         & 50.83                         & 54.81                         & 57.64                          \\
LLaMA-2\textsubscript{13B} & 53.39 & 68.71 & 73.95 & 79.22 & 39.82 & 45.36 & 50.60 & 55.68 & 40.61 & 43.29 & 53.72 & 57.88 \\
GPT-NER & 54.61 & 68.25 & 78.17 & 80.60 & 40.81 & 46.37 & 52.19 & 55.92 & 42.37 & 44.82 & 55.20 & 58.62 \\
\hdashline
\textbf{ABEX-Abs} & 54.18 & 65.52 & 72.36 & 79.40 & 24.62 & 35.28 & 44.71 & 47.90 & 30.76 & 35.26 & 43.28 & 50.60 \\
\textbf{ABEX-stage-2}      & 68.22                         & 71.15                         & 77.02                         & 82.41                          & 41.25                         & 48.73                         & 54.14                         & 54.36                          & 45.85                         & 47.92                         & 55.88                         & 57.62                          \\
\textbf{ABEX-stage-1}                               & {\ul 68.74}                   & 72.09                         & 78.51                         & 83.22                          & 41.28                         & 49.44                         & 54.73                         & 55.60                          & {\ul 46.82}                   & 45.71                         & 56.63                         & 59.25                          \\
\textbf{ABEX} \textit{(ours)} & \cellcolor{magenta!20}\textbf{70.16}                & \cellcolor{magenta!20}\textbf{73.67}                & \cellcolor{magenta!20}\textbf{83.58}                & \cellcolor{magenta!20}\textbf{84.20}                 & \cellcolor{magenta!20}\textbf{43.05}                & \cellcolor{magenta!20}\textbf{51.75}                & \cellcolor{magenta!20}\textbf{56.03}                & \cellcolor{magenta!20}\textbf{58.41}                 & \cellcolor{magenta!20}\textbf{48.76}                & {\ul 51.38}                   & \cellcolor{magenta!20}\textbf{61.85}                & \cellcolor{magenta!20}\textbf{63.14}                 \\
 & $\pm0.86$                & $\pm0.37$                & $\pm1.27$                & $\pm0.31$                 & $\pm0.67$                & $\pm1.32$                & $\pm0.24$                & $\pm1.24$                 & $\pm1.23$                & $\pm0.06$                   & $\pm0.26$                & $\pm0.35$                 \\
 \bottomrule           
\end{tabular}
}}
\caption{\small Result comparison on NER. ABEX outperforms all our baselines by 0.33\% - 36.82\%.}
\label{tab:ner}
\end{table*}

{\noindent \textbf{Gold-only.}} Gold-only refers to training our model only on the low-resource gold training data. 

{\noindent \textbf{SC Baselines.}} For SC, we compare ABEX with text editing baselines: EDA \cite{wei2019eda}, AEDA \cite{karimi-etal-2021-aeda-easier}, and AMR-DA~\cite{shou-etal-2022-amr}, learning-based infilling baselines: SSMBA \cite{ng-etal-2020-ssmba}, GENIUS(-\textbf{ft} version from the original paper) \cite{guo2022genius}, PromDA~\cite{wang-etal-2022-promda}, LLM-based prompting baselines: ZeroGen~\cite{ye-etal-2022-zerogen}, GPT3Mix~\cite{yoo-etal-2021-gpt3mix-leveraging} and rephrasing baselines: BackTrans~\cite{yu2018qanet}. 
\vspace{0.5mm}

{\noindent \textbf{IC Baselines.}} For SC's IC task subset, we add PromptMix~\cite{sahu2023promptmix} as another LLM-based prompting baseline. 
\vspace{0.5mm}

{\noindent \textbf{NER Baselines.}} For NER, we compare with LwTR \cite{dai-adel-2020-analysis}, DAGA \cite{ding-etal-2020-daga}, MulDA \cite{liu2021mulda}, MELM \cite{zhou2021melm} and PromDA~\cite{wang-etal-2022-promda}. 

{\noindent \textbf{QA Baselines.}} For QA, we compare it with ZeroGen, BackTrans, GENIUS, EDA, and AEDA. For SS, we use BackTrans, EDA, AEDA, SSMBA, and AMR-DA. 

{\noindent \textbf{Additional Details.}} For all LLM-based baselines (ZeroGen, GPT3Mix, and PromptMix), we employ LLaMa-13B for a fair comparison. Additionally, for all baselines, we generate 5 synthetic augmentations for a fair comparison. The working of all baselines is detailed in Appendix~\ref{sec:baseline_details}. In all our result tables, ABEX refers to a model trained on synthetic data with optional fine-tuning after training. Finally, we also employ LLaMA-2\textsubscript{13B} as a baseline, where we prompt the LLM to first abstract and then expand. For abstraction, we employ the same prompt in Section~\ref{subsec:learning}. For expansion, we provide the prompt in Appendix~\ref{sec:prompt}.
\vspace{0.5mm}

{\noindent \textbf{Ablations.}} As ABEX ablations, we compare our model with \textbf{ABEX-stage-2}, which does include the fine-tuning on $\mathcal{D}_{ab}$, \textbf{ABEX-stage-1}, which does not include optional fine-tuning on $\mathcal{D}_{down}$ and \textbf{ABEX-Abs}, which does not include the expansion stage and only trains on abstracts as augmentations.

\section{Results and Analysis}
\label{sec:results}

{\noindent \textbf{Quantitative Results.}} Table \ref{tab:classification} compares ABEX on the SC task with our baselines. ABEX outperforms all our baselines by 0.04\% - 29.12\% except on IMDB on the 1000 low-resource setting, where the downstream model overfits the train distribution post data augmentation. Table \ref{tab:ner} compares ABEX on the NER task where ABEX outperforms all our baselines by 0.33\% - 36.82\%. Table \ref{tab:sentence_sim} compares ABEX on the SS task where ABEX outperforms most of our baselines by 0.48\% - 11.22\%. Finally, Table \ref{tab:qa} compares performance on the QA task, where ABEX outperforms all our baselines by 4.05\% - 38.8\%. Text-editing baselines like EDA and LwTR are most competitive to ABEX, while generative ones like DAGA and GENIUS lag behind by considerable margins.
\vspace{0.5mm}

{\noindent \textbf{Robustness against Spurious Correlations.}} Data augmentation methods often amplify spurious correlations in the training set~\cite{evuru2024coda}. ABEX strikes a better balance between consistency and diversity, which would prove to be beneficial in OOD scenarios. Table~\ref{tab:spurious} further compares ABEX performance on SNLI and MNLI with spurious correlations. 
\begin{table}[h!]
\centering
\resizebox{0.5\columnwidth}{!}{%
{\renewcommand{\arraystretch}{0.9}%
\begin{tabular}{l|c|c}
\toprule \toprule
       & SNLI & MNLI \\
    \midrule
Gold-only    &  {\ul80.34}    &  {\ul75.75}   \\
EDA    &   72.68   &   70.90   \\
Genius &   74.64   &   71.26   \\
\textbf{ABEX}~\textit{(ours)}   &   \cellcolor{magenta!20}\textbf{82.88}   &  \cellcolor{magenta!20}\textbf{78.25}  \\
\midrule \bottomrule
\end{tabular}
}}
\caption{\small Result comparison for datasets with known biases.}
\label{tab:spurious}
\end{table}

\begin{table}[t]
\centering
\resizebox{\columnwidth}{!}{%
{\renewcommand{\arraystretch}{1.0}%
\begin{tabular}{l|ccc||ccc}
\toprule \midrule
\textbf{Method}     & \textbf{P($\downarrow$)} & \textbf{D($\uparrow$)} & \textbf{D-L($\uparrow$)} & \textbf{P($\downarrow$)} & \textbf{D($\uparrow$)} & \textbf{D-L($\uparrow$)}\\ \midrule
                    & \multicolumn{3}{c||}{100}                                        & \multicolumn{3}{c}{500}        \\ \midrule
EDA                           & 135.12                            & 103.49                          & 10.63                           & 147.06                             & 120.69                          & 12.07                          \\
SSMBA                         & 86.13                             & 126.66                          & 17.58                            & 103.92                           & 134.44                     & 19.12                         \\
AEDA                          & 105.92                            &  49.72                           & 6.55                        & 106.87                             & 50.56                         & 6.99                           \\
BackTrans                     & 77.17                            & 34.02                          & 19.39                           & 74.98                             & 47.22                         & 20.91                            \\
GPT3-Mix                         & 90.50                            & 124.02                          & 23.55                      & 85.49                             & 134.08                        & 26.98                          \\
GENIUS                        & 32.88                        & {\ul156.50}                    & {\ul27.95}                           & 32.71                             & {\ul159.49}                        & {\ul28.13}                     \\
AMR-DA                        & 68.22                          & 68.73                          & 2.58                         & 64.95                            & 75.15                          & 2.92                            \\
LWTR                          & 152.69                           & 101.95                         & 11.39                          & 137.03                            & 109.02                      & 11.64                           \\
DAGA                          & 66.46                            & 54.59                          & 14.91                            & 120.74                            & 69.32                          & 10.74                        \\
MELM                       & 69.13                             & 113.39                          & 12.91                            & 83.43                             & 116.59                         & 11.30                            \\
\hdashline
\textbf{ABEX-stage-1} \textit{\textit{(ours)}} & {\ul27.46}       & \cellcolor{magenta!20}\textbf{190.87}   & 27.74                     & {\ul26.48}                       & \cellcolor{magenta!20}\textbf{217.29}   &  17.88                         \\
\textbf{ABEX} \textit{\textit{(ours)}} & \cellcolor{magenta!20}\textbf{28.05}       & 124.91   & \cellcolor{magenta!20}\textbf{29.73}                     & \cellcolor{magenta!20}\textbf{27.09}                       & 130.25   &  \cellcolor{magenta!20}\textbf{31.37}                         \\ 
\midrule \bottomrule
\end{tabular}%
}}
\caption{\small Quantitative evaluation of generation quality on the measures of perplexity (P), token diversity (D), and length diversity (D-L). ABEX outperforms all our baselines.
}
\label{tab:qual_table}
\end{table}

\vspace{0.5mm}

\begin{figure*}[t!]
    \centering
    \includegraphics[width=0.98\textwidth]{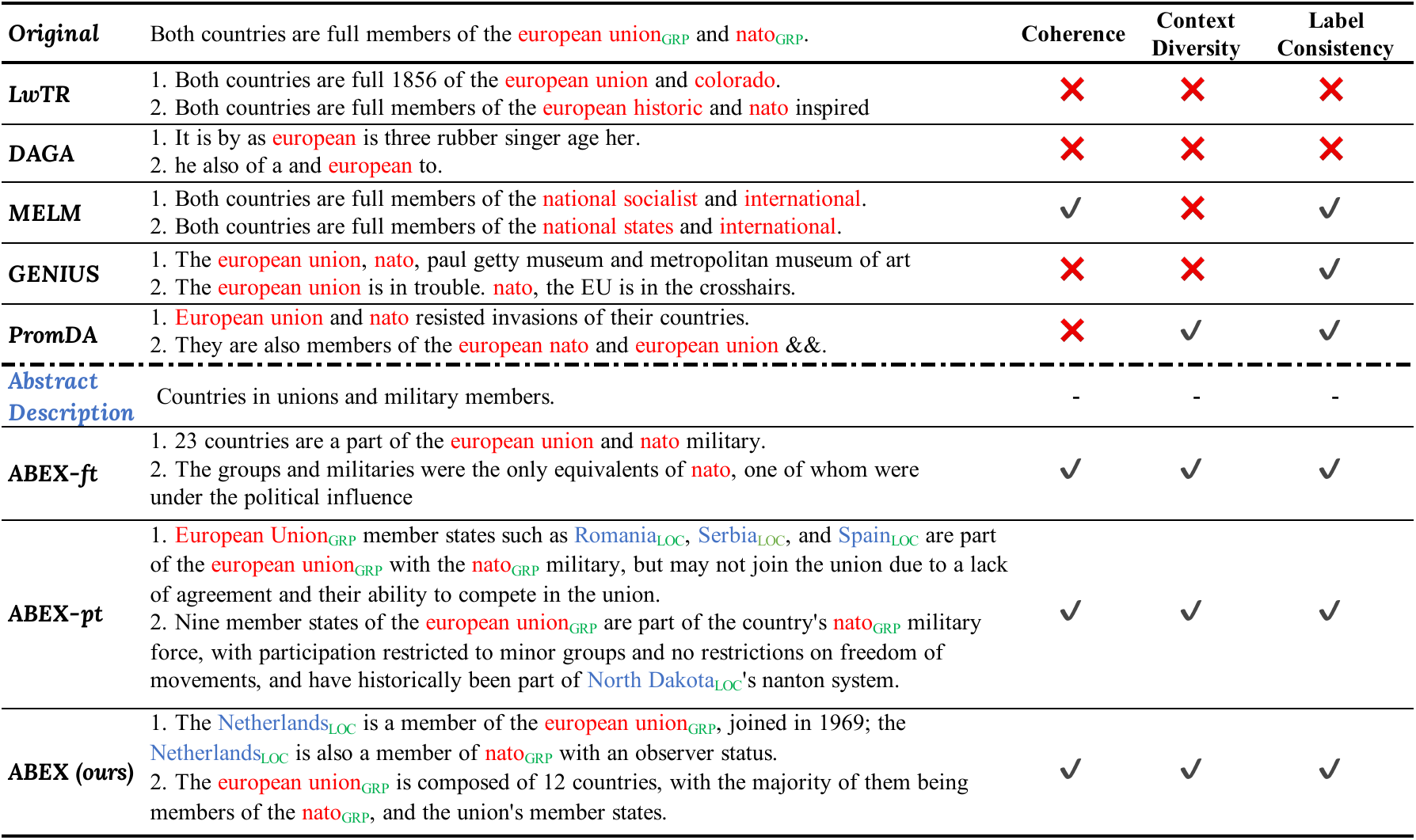}
    \caption{\small{Comparison of augmentations on the MultiCoNER dataset (500 setting). ABEX not only introduces novel contexts of varying lengths around \textcolor{red}{existing NEs} but also introduces \textcolor{blue}{new NEs}. More examples in Fig.~\ref{fig:atis}, \ref{fig:mrpc}, and \ref{fig:yahoo}.}}
    \label{fig:examples_qual}
\end{figure*}

{\noindent \textbf{Qualitative Results.}} Table \ref{tab:qual_table} compares the generation quality of ABEX with all our baselines (averaged baseline-wise across all tasks and splits) on the measures of perplexity \cite{jelinek1977perplexity}, diversity (average percentage of new tokens introduced in $R$ augmentations relative to the total tokens in the original document) and length diversity (average absolute difference in length of source and $R$ augmentations). ABEX outperforms all our baselines in all settings. 

Figure~\ref{fig:examples_qual} compares ABEX augmentations with our baselines on MultiCoNER~\cite{malmasi-etal-2022-multiconer}, a dataset with relatively complex semantics. We define Coherence as the quality of the generated augmentation to be linguistically coherent. We define Label Consistency as the quality of the generated augmentation to maintain the same label as the original sample from which the augmentation was generated. Finally, we define Context Diversity as the quality of the generated augmentation to generate a context around the TRI that is diverse and unique compared to the original document. For all 3 criteria, we provide a red cross if it doesn't meet them and a green tick if it does. ABEX consistently generates augmentations that are coherent, diverse, and label-consistent. The augmentations demonstrate significantly higher degrees of context, entity, and length diversity. Additional examples can be found in Fig.~\ref{fig:atis}, \ref{fig:mrpc}, and \ref{fig:yahoo}, where we also demonstrate that ABEX maintains key syntactic features of the document, such as its style. This is particularly beneficial for tasks like IC, where other methods often alter the style from a question to a statement, negatively impacting performance.


\section{Conclusion}
\label{sec:conclusion}

This paper proposes ABEX, a novel data augmentation framework based on a novel paradigm -- Abstract-and-Expand. Abstract-and-Expand involves first abstracting a given document and then expanding it. To achieve this, we fine-tune BART on a large-scale synthetic dataset to learn expanding abstract descriptions and then propose a controllable and training-free method to generate abstract descriptions for downstream dataset documents by editing AMR graphs. ABEX outperforms all our baselines, quantitatively and qualitatively, on various downstream datasets and tasks.



\section*{Limitations and Future Work}
\label{sec:limitations}
In this section, we list down some potential limitations of ABEX:
\begin{enumerate}
    \item Sentences generated by ABEX may lack factuality. Though factuality is not a requirement for generated synthetic data that serve as augmentations, and most data augmentation methods from literature don't guarantee~\cite{ghosh-etal-2023-dale}, we would like to explore ways to overcome this in future work by methods like knowledge-graph grounded decoding.
    \item Due to its propensity for creating augmentations that are not factually accurate, ABEX is unsuitable for generative tasks such as instruction tuning or generative question answering. Generative natural language understanding (NLU) tasks acquire new knowledge during training, and the introduction of non-factual augmentations by ABEX could negatively impact this knowledge acquisition. The core mechanism of ABEX involves introducing additional augmentations centered around Targeted Reference Information (TRI), which is beneficial primarily for discriminative tasks like sequence classification, named entity recognition (NER), question answering (QA), and others. This is because the model in these tasks focuses on identifying patterns in the data rather than acquiring new information. The introduction of varied contexts by ABEX enhances the model's ability to learn these discriminative patterns more efficiently and adapt to new, unseen data distributions. Consequently, in alignment with previous methodologies, our evaluation of ABEX is limited to discriminative NLU tasks, excluding generative tasks.
    \item ABEX depends on pre-trained AMR-to-Text and Text-to-AMR models for controllable abstract generation. However, AMR parsing is not a solved problem; these models often make errors. Therefore, as part of future work, we would like to explore better methods for controllable abstract generation.
    
\end{enumerate}


\bibliography{anthology,custom}

\appendix
\section{Hyper-parameter Tuning}
\label{sec:hyper-parameter}


\subsection{Effect of $\mu$ on the diversity of generations}
\label{sec:p_gen}

Table \ref{tab:mu-diversity} compares the performance and the diversity of augmentations generated by ABEX at different values of $\mu$. The parameter $\mu$ plays a crucial role in controlling the deletion rate $\varepsilon$ during the editing of the AMR graph. By increasing the mean of the Gaussian distribution, we observe a corresponding increase in the average deletion rate, leading to a higher level of abstraction. Consequently, this strategy enhances the performance and diversity of generated augmentations, reaching a peak value before exhibiting a decline.

\begin{table}[h]
    \centering
    \small
    \resizebox{\columnwidth}{!}{
    \begin{tabular}{l|ccccccc}
    \hline
         \textbf{$\mu$} & \textbf{0.2} & \textbf{0.3} & \textbf{0.4} & \textbf{0.5} & \textbf{0.6} & \textbf{0.7}\\
        \hline
          $F_1$& 65.41 & 65.76 & 67.83 & \cellcolor{magenta!20}\textbf{69.99} & 67.60 & 67.37  \\
          Diversity & 192.73 & 195.61 & 198.27 & \cellcolor{magenta!20}\textbf{201.63} & 195.76 & 193.28\\
          Diversity-L & 28.09 & 28.82 & 29.33 & \cellcolor{magenta!20}\textbf{30.17} & 29.63 & 28.29\\
         \hline
    \end{tabular}
    }
    \caption{F1 and diversity metrics for various settings of $\mu$. All values are averaged across all datasets for all low-resource settings.}
    \label{tab:mu-diversity}
\end{table}

\subsection{Effect of augmentation rounds $R$}
\label{sec:augrounds}

Table \ref{tab:aug} compares the performance of ABEX at different values of $R$. Augmenting the training dataset with several augmentation rounds $R$ proves effective until the model overfits to the training data. The observation is similar to prior work in data augmentation for NLU tasks \cite{zhou2021melm,ghosh-2023-aclm}.

\begin{table}[h]
    \centering
    \small
    \resizebox{\columnwidth}{!}{
    \begin{tabular}{c|ccccccc}
    \hline
         \textbf{$R$} & \textbf{1} & \textbf{2} & \textbf{3} & \textbf{4} & \textbf{5} & \textbf{6} & \textbf{7}\\
        \hline
          $F_1$& 67.65 & 67.99 & 69.06 & 69.64 & \cellcolor{magenta!20}\textbf{69.99} & 69.71 & 69.22  \\
         \hline
    \end{tabular}
    }
    \caption{F1 for various settings of $R$. All values are averaged across all datasets for all low-resource settings.}
    \label{tab:aug}
\end{table}

\subsection{Effect of $\alpha$}
\label{sec:alpha}

Table \ref{tab:alpha-tab} compares the performance of ABEX at different values of $\alpha$. While a lower $\alpha$ leads to deleting smaller sub-graphs which would effectively decrease abstraction, a higher $\alpha$ leads to deleting bigger sub-graphs and thus higher abstraction. Similar to our finding in Section \ref{sec:p_gen}, training and inferring with highly abstract sentences leads the model to generate sentences that do not match the underlying data distribution and, thus, sub-optimal performance.

\begin{table}[h]
\centering
\resizebox{0.8\columnwidth}{!}{%
\begin{tabular}{c|cccccc}
\hline
$\alpha$  & \textbf{0.25} & \textbf{0.30} & \textbf{0.35}  & \textbf{0.40} & \textbf{0.45} & \textbf{0.5} \\ \hline
 $F_1$&  65.63    &   68.89   & \cellcolor{magenta!20}\textbf{69.99} &  69.97    &  68.11    &  68.90  \\ \hline
\end{tabular}%
}
\caption{F1 for various settings of $\alpha$. All values are averaged across all datasets and all low-resource settings.}
\label{tab:alpha-tab}
\end{table}

\subsection{Effect of $\beta$}
\label{sec:beta}
Table \ref{tab:beta-tab} compares the performance of ABEX augmentations at different values of $\beta$. A lower $\beta$ leads to less diverse sentences (as a result of lesser augmentations generated using mixed abstracts), and a higher $\beta$ leads to more diverse sentences (as a result of more sentences generated using mixed abstracts). While token diversity in augmentations improves performance, too much might lead to sub-optimal performance.

\begin{table}[h]
\centering
\small
\resizebox{0.8\columnwidth}{!}{%
\begin{tabular}{c|cccccc}
\hline
$\beta $  & \textbf{0} & \textbf{0.2} & \textbf{0.4}  & \textbf{0.6} & \textbf{0.8} & \textbf{1} \\ \hline
 $F_1$&  69.90    &  69.77    & 69.93 & \cellcolor{magenta!20}\textbf{69.99}      &  68.86    &  68.21  \\ \hline
\end{tabular}%
}
\caption{F1 for various settings of $\beta$. All values are averaged across all datasets and all low-resource settings.}
\label{tab:beta-tab}
\end{table}


\section{Prompts}
\label{sec:prompt}

{\noindent \textbf{Document - to - Summary}} For summarizing a document from $\mathcal{D}_u$ with LLaMA-2, we use the following prompt: \textit{Write me a summary of the article in one line. Don't include entities; write the summary just describing key events and concepts in the article. Here is the article:}.
\vspace{0.5mm}

{\noindent \textbf{Summary - to - Abstract}} For generating an abstract from the summary of a document in $\mathcal{D}_u$ with LLaMA-2 we use the following prompt: \textit{I will provide you with a small document. You need to return a short and abstract description of it. Don't mention named entities, and just describe the key message of the document in a few words.
Here are some examples:
Input 1: Shatrughan Sinha, a Congress candidate and actor-politician, will run against Union Law Minister Ravi Shankar Prasad, a BJP candidate, in the Patna Sahib seat. Sinha has dismissed BJP's claim that the seat is their stronghold and has expressed his confidence in winning the election. He has also criticized the BJP's decision to field Prasad, a four-term Rajya Sabha member, in the seat. Sinha has served two terms in the Rajya Sabha and has been a member of the union council of ministers. He has also defended his record, citing his spending of 106\% of his MPLAD fund, which is available on the net.
Output 1: A political competition between two candidates from major parties for a significant electoral seat, involving critique of the opposition's choice and defense of personal achievements.
Input 2: Said Baalbaki, a Palestinian artist, has curated an exhibition featuring 50 of Abbo's sketches, etchings, and objects, along with texts from Baalbaki's personal collection, showcasing the elusive sculptor's work and life.
Output 2: An exhibition curated by an artist, displaying sketches, etchings, and objects from a lesser-known sculptor, accompanied by personal texts, highlighting the sculptor's work and life.
Here is the input document:}. The exemplars are human written.

{\noindent \textbf{Abstract - to - Expansion (for LLaMA-13B baseline)}} \textit{I will provide you with an abstract version of a document. You need to understand the abstract and return an expanded version of the document from the abstract. The expansion can be diverse and can add new context and entities. However, it should follow the following constraints while expanding: 1) It should be semantically similar to the abstract, i.e., retain the key points and the message in the abstract. 2) It should retain the following keywords or phrases: [TRI extracted from Section~\ref{subsec:learning}] 3) The generated sentence should be of the label [Ground-truth document label]. Here is an example of a sentence from the label [Randomly retrieved sentence with the same label]. Here are some examples: [2 Human written exemplars of the process]}

\section{Algorithm}
\label{sec:algorithm}

We show the Algorithm for ABEX in Algorithm \ref{alg:algo1}.


\section{Dataset Details}
\label{sec:dataset_details}

\begin{table*}[t]
    \centering
    \resizebox{\textwidth}{!}{
    \begin{tabular}{llclcc}
    \toprule
         \textbf{Dataset} & \textbf{Source} & \textbf{Sub-domain} & \textbf{Task Type} & \textbf{Training/Dev/Test Instances} & \textbf{Classes} \\
         \midrule
          HuffPost &\citet{huffpost} & HuffPost website & Multi-class classification & 67490/16891/16891 & 5\\
         Yahoo & \citet{zhang2015character} & Yahoo Answers  & Multi-class classification & 1375404/58966/58966 & 10  \\
          IMDB & \cite{maas-EtAl:2011:ACL-HLT2011} & IMDB Reviews & Multi-class classification & 25000/-/25000 & 2  \\
         CoNLL-2003 & \citet{tjong-kim-sang-de-meulder-2003-introduction}  & English news articles & Named Entity Recognition & 14041/3250/3453 & 4 \\
         MultiCoNER  & \citet{malmasi-etal-2022-multiconer} & Search Queries & Named Entity Recognition & 15300/800/217818 & 6 \\
         OntoNotes-5.0 & \citet{pradhan2013towards} & Diverse & Named Entity Recognition & 115812/15680/12217 & 36 \\
         ATIS & \citet{CNTK_2023} & Travel enquiry & Intent Classification & 4972/888/888 & 17 \\
         MASSIVE & \citet{fitzgerald2022massive} & Multidomain & Intent Classification & 11500/2030/2970 & 60 \\
         MRPC & \citet{dolan2005automatically} & English news articles & Sentence Similarity & 3668/408/1725 & 2 \\
         QQP & \citet{quora-question-pairs} & Quora questions & Sentence Similarity  & 363846/40430/40430 & 2 \\
         SQuAD & \citet{rajpurkar2016squad} & Wikipedia Articles & Question Answering & 87600/10600/- & - \\
         NewsQA & \citet{trischler-etal-2017-newsqa} & CNN Articles & Question Answering & 92549/5126/5166 & - \\
         SNLI & \cite{bowman-etal-2015-large} & Human Written Sentences & Natural Language Inference & 550000/10000/- & 3 \\
         MNLI & \cite{williams-etal-2018-broad} & CNN Articles & Question Answering & 393000/19650/- & 3 \\
         \bottomrule
    \end{tabular}
    }
    \vspace{-2mm}
    \caption{Statistics for each downstream NLU datasets used in our experiments. As described in Section \ref{subsec:task_dataset}, we derive low-resource splits from these original datasets for our experiments.}
    \label{tab:nlu_datasets}
    \vspace{-5mm}
\end{table*}

\subsection{Classification}
\label{subsec:classification}

{\noindent \textbf{HuffPost.}}  The HuffPost dataset \cite{huffpost} is a popular multiclass classification dataset in NLP. It is a collection of news articles from the HuffPost website, covering a wide range of topics, including politics, business, entertainment, and more. For multiclass classification, the HuffPost dataset is labeled with a diverse set of categories and for our experiments, we take sentences from five categories, including politics, sports, entertainment, tech, and business. Dataset statistics can be found in Table \ref{tab:nlu_datasets}.
\vspace{0.5mm}

{\noindent \textbf{Yahoo.}} The Yahoo Answers topic classification dataset \cite{zhang2015character} is a widely used dataset for multi-class text classification tasks. It is derived from the Yahoo Answers community-driven question-answering platform, where users ask questions on various topics, and community members provide answers. The dataset contains a large number of question-and-answer pairs covering a wide range of categories or topics. Each question in the dataset is associated with one primary category. The primary categories span diverse subjects, including Society \& Culture, Science \& Mathematics, Health, Education \& Reference, Computers \& Internet, Sports, Business \& Finance, Entertainment \& Music, Family \& Relationships, Politics \& Government, Travel, Cars \& Transportation, Food \& Drink, Games \& Recreation, Home \& Garden, Local Businesses, News \& Events, Pets, Beauty \& Style and Pregnancy \& Parenting. Dataset statistics can be found in Table \ref{tab:nlu_datasets}.
\vspace{0.5mm}

\algdef{SE}[FOR]{NoDoFor}{EndFor}[1]{\algorithmicfor\ #1}{\algorithmicend\ \algorithmicfor}
\algdef{SE}[IF]{NoThenIf}{EndIf}[1]{\algorithmicif\ #1}{\algorithmicend\ \algorithmicif}
\renewcommand{\algorithmiccomment}[1]{\hfill$\triangleright${#1}}
\begin{algorithm}[h]
\scriptsize
\caption{ABEX: Our proposed augmentation framework}
\label{alg:algo1}
    \begin{algorithmic}
    \State \textbf{ABEX Pre-training}
    \State Given an instruction-tuned LLM, unlabelled dataset $\mathrm{D}_{u}$, and pre-trained BART
    \State \text{Synthesize } $\mathrm{D}_{ab}$ \text{ with abstract-document pairs by prompting the LLM on } $\mathrm{D}_{u}$
    \State $\text{Train BART on } \mathrm{D}_{u}$
     \State \textbf{Data Augmentation with pre-trained BART}
    \State $\textbf{\text{Given }} \text{training set } \mathbb{D}_{\text{down}}, \text{ and pre-trained BART on } \mathbb{D}_{\text{u}}$
    \State $\mathbb{D}_{ab} \gets \emptyset, \mathbb{D}_{aug} \gets \emptyset$
    \NoDoFor {${\{X,Y\}} \in \mathbb{D}_{train}$} \textbf{do}
    \Comment{Training Loop}
    \State $t_{amr} \gets \textsc{TextToAmr}(X)$
    \State $t^{'}_{amr} \gets \textsc{FilterAttr}(t_{amr})$
    \Comment{Remove Attributes}
    \State $t^{'}_{amr} \gets \textsc{DeleteSubTree}(t^{'}_{amr}), \text{ if depth-ratio < }\alpha$
    \State $\tilde{X} \gets \textsc{AmrToText}(t^{'}_{amr})$
    \State $\mathbb{D}_{\text{abstract}} \leftarrow \mathbb{D}_{\text{abstract }} \cup\{\tilde{X},Y\}$
    \EndFor
    \NoDoFor {${\{\tilde{X},Y\}} \in \mathbb{D}_{abstract}$} \textbf{do}
    \State $\textsc{BART}_{finetune } \leftarrow \textsc{Finetune}(\textsc{BART}, \tilde{X})$
    \Comment{Fine-tune BART}
    \EndFor
    \NoDoFor $ \{{{X,Y\}} \in \mathbb{D}_{down}} \textbf{\text{ do}}$
    \Comment{Generation Loop}
    \State $\textbf{\text{repeat }} \mathcal{R} \textbf{\text{ times}}$:
            \State $t_{amr} \gets \textsc{TextToAmr}(X)$
            \State $t^{'}_{amr} \gets \textsc{FilterAttr}(t_{amr})$
            \Comment{Remove Attributes}
            \State $t^{'}_{amr} \gets \textsc{DeleteSubTree}(t^{'}_{amr}), \text{ if depth-ratio < }\alpha$
            \State $X^{'} \gets \textsc{Similar}(X)$
            \Comment{Semantically similar sentence}
            \State $ST \gets \textsc{SubTreePairs}(X, X^{'})$
            \State $\forall \text{ } (x_1, x_2) \in ST,$
            \State $t_{sim} \gets \textsc{ArgMax} (\textsc{Smatch++}(x_1, x_2))$
            \State $t^{'}_{mix} = t^{'}_{amr} + t_{sim}$
            \Comment{Append similar subtree}
            \State $\tilde{X} \gets \textsc{AmrToText}(t^{'}_{amr})$
            \State $\tilde{X}_{mix} \gets \textsc{AmrToText}(t^{'}_{mix})$
            \State $X_{aug} \leftarrow \textsc{BART}_{finetune}(\tilde{X}), \text{ } if \text{ } \gamma < \beta$
            \State $X_{mix} \leftarrow \textsc{BART}_{finetune}(\tilde{X}_{mix}), \text{ } if \text{ } \gamma > \beta$
            \State $\mathbb{D}_{aug } \leftarrow \mathbb{D}_{aug} \cup \{X_{aug},Y\}\cup \{X_{mix},Y\}$
    \EndFor
    \State $\mathbb{D}_{aug} \leftarrow \textsc{PostProcess}(\mathbb{D}_{aug})$
    \Comment{Post-processing}
    \State \textbf{return} $\mathbb{D}_{train} \cup \mathbb{D}_{aug}$
    \end{algorithmic}
\end{algorithm}

\subsection{Named Entity Recognition}
\label{subsec:ner}

{\noindent \textbf{CoNLL-2003.}} The CoNLL-2003 dataset \cite{tjong-kim-sang-de-meulder-2003-introduction} is a widely used benchmark dataset for Named Entity Recognition (NER) tasks in NLP. It was created for the Conference on Computational Natural Language Learning (CoNLL) shared task in 2003. The dataset consists of news articles from the Reuters Corpus, a collection of English news articles. It is annotated with four named entities: person, organization, location, and miscellaneous entities (such as dates and percentages). The annotations indicate the boundaries of the named entities within the text. Dataset statistics can be found in Table \ref{tab:nlu_datasets}.
\vspace{0.5mm}

{\noindent \textbf{MultiCoNER.}} MultiCoNER  \cite{malmasi-etal-2022-multiconer} is large multilingual dataset for complex NER. MultiCoNER covers 3 domains, including Wiki sentences, questions, and search queries, across 11 distinct languages. The dataset represents contemporary challenges in NER and is labeled with six distinct types of entities: \textbf{person}, \textbf{location}, \textbf{corporation}, \textbf{groups} (political party names such as \emph{indian national congress}), \textbf{product} (consumer products such as \emph{apple iPhone 6}), and \textbf{creative work} (movie/song/book titles such as \emph{on the beach}). Dataset statistics can be found in Table \ref{tab:nlu_datasets}.
\vspace{0.5mm}

{\noindent \textbf{Ontonotes 5.0.}} Ontonotes 5.0 \citet{pradhan2013towards} is a widely used dataset in the field of Natural Language Processing (NLP) and specifically for Named Entity Recognition (NER) tasks. It is a large-scale corpus that provides annotations for a variety of linguistic phenomena, including named entities, across multiple languages. The dataset contains a diverse range of text genres, including news articles, conversational data, and web data, making it suitable for training and evaluating NER models in different domains. It covers three languages: English, Chinese, and Arabic. The dataset is annotated with 11 categories: Person, Organization, Location, Date, Time, Money, Percent, Quantity, Ordinal and Miscellaneous. Dataset statistics can be found in Table \ref{tab:nlu_datasets}.

\subsection{Intent Classification}
\label{subsec:intent}
{\noindent \textbf{ATIS.}} The ATIS (Airline Travel Information System) dataset\footnote{\url{https://github.com/howl-anderson/ATIS_dataset/tree/master}} is a widely used benchmark dataset for intent classification in the field of NLU. It was developed to address understanding user intents in the context of airline travel information. The dataset consists of queries or utterances that users might input when interacting with a flight reservation system. Each query is labeled with an intent representing the user's intention or purpose behind the query. 
The dataset is labeled with intents that are: Flight-Booking, Flight-Status, Flight-Information, Ground-Service, Airfare, Airport-Information, Travel-Preferences, Flight-Cancellation, and None/No-Intent. Dataset statistics can be found in Table \ref{tab:nlu_datasets}.
\vspace{0.5mm}

{\noindent \textbf{MASSIVE.}} The MASSIVE (Multilingual Amazon Slu resource package for Slot-filling) \citet{fitzgerald2022massive} dataset is a widely used benchmark dataset for intent classification in the field of NLU. It contains 1M realistic, parallel, labeled virtual assistant utterances spanning 51 languages, 18 domains, 60 intents, and 55 slots. The dataset is labeled with intents some of which are: Alarm set, Play music, Audio volume mute, Weather query, Takeaway order and General joke etc. Dataset statistics can be found in Table \ref{tab:nlu_datasets}.

\subsection{Sentence Similarity}
\label{subsec:sentence}

{\noindent \textbf{MRPC.}} The Microsoft Research Paraphrase Corpus (MRPC) dataset \cite{dolan2005automatically} is a benchmark for paraphrase identification and semantic similarity tasks. It was developed by Microsoft Research to support research in natural language processing (NLP) and machine learning. The MRPC dataset consists of pairs of sentences manually annotated as either paraphrases (sentences with similar meanings) or non-paraphrases (sentences with different meanings). The sentences cover various domains and topics, including news, fiction, and general web data. Dataset statistics can be found in Table \ref{tab:nlu_datasets}.
\vspace{0.5mm}

{\noindent \textbf{QQP.}} The Quora Question Pairs (QQP) dataset\footnote{\url{https://quoradata.quora.com/First-Quora-Dataset-Release-Question-Pairs}} is a widely used benchmark dataset in the field of natural language processing (NLP). It was created by Quora, a popular question-and-answer platform, and released for research. The QQP dataset consists of pairs of questions collected from the Quora platform. Each question pair is labeled as duplicate or non-duplicate, indicating whether the two questions have the same meaning. The dataset contains many question pairs covering diverse topics, allowing for the exploration of semantic similarity and question-matching tasks. Dataset statistics can be found in Table \ref{tab:nlu_datasets}.

\subsection{Question Answering}
\label{subsec:qa}
{\noindent \textbf{SQUAD.}} The SQUAD (Stanford Question Answering Dataset) \cite{rajpurkar2016squad} is a reading comprehension dataset, consisting of questions posed by crowdworkers on a set of Wikipedia articles, where the answer to every question is a segment of text, or span, from the corresponding reading passage, or the question might be unanswerable. Dataset statistics can be found in Table \ref{tab:nlu_datasets}.
\vspace{0.5mm}

{\noindent \textbf{NEWSQA.}} NewsQA (News Question Answering) \cite{trischler-etal-2017-newsqa} is a challenging machine comprehension dataset of over 100,000 human-generated question-answer pairs. Crowdworkers supply questions and answers based on a set of over 10,000 news articles from CNN, with answers consisting of spans of text from the corresponding articles. Dataset statistics can be found in Table \ref{tab:nlu_datasets}.

\subsection{Bias Testing}
\label{subsec:bias}
{\noindent \textbf{SNLI.}} The SNLI (Stanford Natural Language Inference) \cite{bowman-etal-2015-large} corpus is a collection of 570k human-written English sentence pairs manually labeled for balanced classification with the labels entailment, contradiction, and neutral, supporting the task of natural language inference (NLI), also known as recognizing textual entailment (RTE). Dataset statistics can be found in Table \ref{tab:nlu_datasets}.
\vspace{0.5mm}

{\noindent \textbf{MNLI.}} The MNLI (Multi-Genre Natural Language Inference) \cite{williams-etal-2018-broad} corpus is a crowd-sourced collection of 433k sentence pairs annotated with textual entailment information. The corpus covers a range of genres of spoken and written text, and supports a distinctive cross-genre generalization evaluation. Dataset statistics can be found in Table \ref{tab:nlu_datasets}.
\vspace{0.5mm}

\section{Baseline Details}
\label{sec:baseline_details}


{\noindent \textbf{SSMBA.}} SSMBA \cite{ng-etal-2020-ssmba} generates synthetic training examples by using a pair of corruption and reconstruction functions to move randomly on a data manifold.
\vspace{0.5mm}

{\noindent \textbf{AEDA.}} AEDA \cite{karimi-etal-2021-aeda-easier} is similar to EDA but only employs random insertion of punctuation marks in the original text to generate synthetic augmentations.
\vspace{0.5mm}

{\noindent \textbf{GENIUS.}} GENIUS \cite{guo2022genius}, pre-trains and optionally fine-tunes BART \cite{lewis2019bart} on a denoising objective using sketches generated with an extreme masking algorithm. The extreme masking algorithm just preserves keywords in a sentence and masks everything else. 
\vspace{0.5mm}

{\noindent \textbf{MELM.}} MELM \cite{zhou2021melm}, which stands for Masked Entity Language Modeling, suggests the fine-tuning of a transformer-encoder-based PLM on linearized labeled sequences through masked language modeling. In low-resource scenarios, MELM surpasses all other baselines and prior techniques on the CoNLL 2003 NER dataset across four languages, including mono-lingual, cross-lingual, and multi-lingual settings.
\vspace{0.5mm}

{\noindent \textbf{DAGA.}} DAGA \cite{ding-etal-2020-daga}, short for Data Augmentation with a Generation Approach, suggests the training of a one-layer LSTM-based recurrent neural network language model (RNNLM) by maximizing the probability of predicting the next token using linearized sentences. For sentence generation, they employ random sampling to create entirely new sentences, with the model being fed only the $\lbrack\textbf{BOS}\rbrack$ token.
\vspace{0.5mm}


{\noindent \textbf{LwTR.}} LwTR \cite{dai-adel-2020-analysis} replaces a token in a sentence with another token of the same label; the token is randomly selected from the training set.

{\noindent \textbf{PromDA.}} PromDA \cite{wang-etal-2022-promda} proposes a data augmentation framework based on T5 that trains soft prompts using a novel keyword-to-sentence algorithm.

{\noindent \textbf{AMR-DA.}} AMR-DA~\cite{shou-etal-2022-amr} converts a sample document from a dataset to an AMR graph, modifies the graph according to various data augmentation policies, and then generates augmentations from graphs. The method combines both sentence-level techniques like back translation and token-level techniques like EDA.

{\noindent \textbf{PromptMix.}} PromptMix~\cite{sahu2023promptmix} PromptMix prompts instruction-tuned LLMs to generate augmentations for text classification tasks that are close to the class boundary.

{\noindent \textbf{ZeroGen.}} ZeroGen~\cite{ye-etal-2022-zerogen}, similar to PromptMix, generates data using LLMs but in a zero-shot manner without any Gold-only  data. It prompts pre-trained LLMs (not instruction fine-tuned) for data synthesis.

{\noindent \textbf{Baselines not considered.}} We do not consider more recent baselines provided by \citet{cai-etal-2023-graph}, \citet{hu2023entitytotext} and \citet{rahamim-etal-2023-text} as the code for the same was not available at the time of writing the paper. Additionally, we do not consider \citet{zhou-etal-2022-flipda} as label flipping is not applicable for our paper for all tasks considered, and \citet{chen-etal-2022-style} as style transfer is better suited for cross-domain tasks and applying it to single domain tasks is not trivial. Finally, we do not consider \citet{yu2023large} as it requires manual human intervention for attribute extraction for a dataset.


\section{Additional Details}
\label{sec:additional}

\subsection{AMR Attributes}
\label{subsec:amr_attributes}

In Section \ref{subsec:sentence_to_abstract}, we describe the removal of a predefined set of attributes from the AMR graph. These sentence-specific attributes are deemed non-essential to the underlying semantics of the sentence and are thus removed. The targeted attributes for removal include: \textbf{:mod}, \textbf{:wiki}, \textbf{:quant}, \textbf{:value} and \textbf{:op}. This process ensures that the resulting AMR graph primarily captures the essential semantic information relevant to the sentence, improving the clarity and conciseness of the abstract description.

\subsection{Similar Sentence Retrieval}
\label{subsec:similar_retreival}

We employ semantic retrieval to mix AMR graphs of 2 semantically similar sentences and generate a single abstract description covering the contents of both sentences. Note that the retrieval uses the original sentence, not the AMR graph of the sentence. Specifically, we calculate the cosine similarity $\mathrm{sim(.)}$ between embeddings $\mathrm{e(a)}$ and  $\mathrm{e(b)}$ as follows:

\begin{equation}
\label{eqtn:cosine}
    \mathrm{sim(a,b)} =\frac{\mathrm{e(a) \cdot e(b)}}{\left\|\mathrm{e(a)}\right\|\left\|\mathrm{e(b)}\right\|}
\end{equation}

where $\mathrm{e(.)}$ is a sentence-encoder (Sentence-BERT in our case) and $a$, and $b$ are text sentences. We take $b$ as the corpus sentence with the highest cosine similarity to $a$.

\subsection{SMATCH++}
\label{subsec:smatch}

SMATCH (Semantic Matching of Nodes Anchored on Trees) is a graph-matching algorithm designed to evaluate the semantic similarity between structured data, such as parse trees or semantic graphs. It is commonly used in NLP and information retrieval tasks. The SMATCH algorithm considers two input graphs and measures their similarity based on the common structure and semantic alignment between nodes. It operates by recursively matching nodes in a top-down manner, considering both the nodes' syntactic relationships and semantic properties. The key idea behind SMATCH is to find the best alignment between nodes of the two input graphs, aiming to maximize the matching score while minimizing structural and semantic inconsistencies. It assigns similarity scores to matched nodes based on their attribute values and relationships and calculates the overall graph similarity as the weighted average of node similarity scores. 


The output of the SMATCH algorithm is a similarity score that quantifies the semantic similarity between the two input graphs. Higher scores indicate greater similarity, while lower scores indicate dissimilarity.

SMATCH aims to measure the structural similarity of graphs via the number of triples shared by $\mathcal{G}_\mathcal{A}$ and $\mathcal{G}_\mathcal{B}$. To obtain a meaningful score, it leverages an alignment $map$:$~vars(a) \leftrightarrow vars(b)$ that tells it how to map a variable in the first MR to a variable in the second MR. In this alignment, at maximum, every variable from $a$ can have one partner in $b$ (and vice versa). Let an application of a $map$ to a graph $a$ be denoted as $a^{map}:=\{t^{map}~;~ t \in a\}$, where  $t^{map}$ of a triple $t = \triple{x}{:rel}{y}$ is set to  $t^{map} = \triple{map(x)}{:rel}{map(y)}$ for binary triples, and $t^{map} = \triple{map(x)}{:rel}{c}$ for unary triples. Under any alignment $map$, we can calculate an overlap score $f$. In original smatch, $f$ is the size of the triple overlap of $a$ and $b$:
\begin{equation}
\label{eq:hardscore}
f(a, b, map) = |a^{map} \cap b|.
\end{equation},

The primary aim is to find $F$ as follows: 

\begin{equation}
\label{eq:smatch}
     F = \max_{map} f(a, b, map),
\end{equation}

Finding a maximizer $map^\star$  lies at the heart of SMATCH. For now, we assume that we have $map^\star$ at our disposal. Therefore, we can calculate \textit{precision} ($P$) and \textit{recall} ($R$):

\begin{align}
\label{eq:pr}
    P = |a|^{-1} F,~~~~~~~R = |b|^{-1} F,
\end{align}
to obtain a final F1 evaluation score: $2PR/(P+R)$. With such a score, we can assess the similarity of MRs, and compare and select parsing systems. 
\vspace{1mm}

SMATCH++ \cite{opitz-2023-smatch} improves over SMATCH by proposing a
standardized and extended metric calculation of fine-grained sub-graph meaning aspects, making it more suitable for our task. Specifically, they show the feasibility of optimal alignment in a standard evaluation setup and develop a lossless graph compression method that shrinks the
search space and significantly increases efficiency. We request our readers to refer to the original paper for more details.

\section{Extra Details}
\label{sec:extra_details}

{\noindent \textbf{Model Parameters:}} BART\textsubscript{large} $\approx$ has 680M parameters with 12 layers of encoder, 12 layers of decoder, 1024-hidden-state, and 16-heads. BERT\textsubscript{base} has $\approx$~110M 12-layers of encoder, 768-hidden-state, 2048 feed-forward hidden-state, and 8-heads.
\vspace{1mm}

{\noindent \textbf{Compute Infrastructure:}} All our experiments are conducted on a single NVIDIA A100 GPU. An entire ABEX training pipeline takes $\approx$ 2 hours.
\vspace{1mm}

{\noindent \textbf{Implementation Software and Packages:}} We implement all our models in PyTorch \footnote{\url{https://pytorch.org/}} and use the HuggingFace \footnote{\url{https://huggingface.co/}} implementations of BERT\textsubscript{base} and BART\textsubscript{large}.

We also use the following repositories for running the baselines: BackTrans~\cite{yu2018qanet}, EDA\footnote{\href{https://github.com/jasonwei20/eda_nlp}{https://github.com/jasonwei20/eda\_nlp}}\cite{wei2019eda}, AEDA\footnote{\href{https://github.com/akkarimi/aeda_nlp}{https://github.com/akkarimi/aeda\_nlp}}~\cite{karimi-etal-2021-aeda-easier}, AMR-DA\footnote{\href{https://github.com/zzshou/amr-data-augmentation}{https://github.com/zzshou/amr-data-augmentation}}~\cite{shou-etal-2022-amr}, SSMBA\footnote{\href{https://github.com/nng555/ssmba}{https://github.com/nng555/ssmba}}~\cite{ng-etal-2020-ssmba}, GENIUS(-\textbf{ft})\footnote{\href{https://github.com/beyondguo/genius}{https://github.com/beyondguo/genius}} \cite{guo2022genius}, PromDA\footnote{\href{https://github.com/GaryYufei/PromDA}{https://github.com/GaryYufei/PromDA}}~\cite{wang-etal-2022-promda}, PromptMix\footnote{\href{https://github.com/servicenow/promptmix-emnlp-2023}{https://github.com/servicenow/promptmix-emnlp-2023}}~\cite{sahu2023promptmix}, ZeroGen\footnote{\href{https://github.com/jiacheng-ye/ZeroGen}{https://github.com/jiacheng-ye/ZeroGen}}~\cite{ye-etal-2022-zerogen}, GPT3Mix\footnote{\href{https://github.com/naver-ai/hypermix}{https://github.com/naver-ai/hypermix}}~\cite{yoo-etal-2021-gpt3mix-leveraging}, LwTR\footnote{\href{https://github.com/boschresearch/data-augmentation-coling2020}{https://github.com/boschresearch/data-augmentation-coling2020}}~\cite{dai-adel-2020-analysis}, DAGA\footnote{\href{https://github.com/ntunlp/daga}{https://github.com/ntunlp/daga}} \cite{ding-etal-2020-daga}\cite{ding-etal-2020-daga} and MELM\footnote{\href{https://github.com/randyzhouran/melm}{https://github.com/randyzhouran/melm}}~\cite{zhou2021melm}. All the baseline repositories are covered under the MIT License.

We use the following datasets to evaluate: Huffpost\footnote{\href{https://www.kaggle.com/datasets/rmisra/news-category-dataset}{https://www.kaggle.com/datasets/rmisra/news-category-dataset}}~\cite{huffpost}, Yahoo\footnote{\href{https://huggingface.co/datasets/yahoo_answers_topics}{https://huggingface.co/datasets/yahoo\_answers\_topics}}~\cite{zhang2015character}, IMDB\footnote{\href{https://ai.stanford.edu/~amaas/data/sentiment/}{https://ai.stanford.edu/~amaas/data/sentiment/}}~\cite{maas-EtAl:2011:ACL-HLT2011}, Massive\footnote{\href{https://huggingface.co/datasets/AmazonScience/massive/viewer/en-US}{https://huggingface.co/datasets/AmazonScience/massive/viewer/en-US}}~\cite{fitzgerald2022massive}, ATIS\footnote{\href{https://github.com/howl-anderson/ATIS_dataset}{https://github.com/howl-anderson/ATIS\_dataset}}~\cite{coucke2018snips}, ConLL-2003\footnote{\href{https://huggingface.co/datasets/conll2003}{https://huggingface.co/datasets/conll2003}}~\cite{tjong-kim-sang-de-meulder-2003-introduction},  OntoNotes-5.0\footnote{\href{https://catalog.ldc.upenn.edu/LDC2013T19}{https://catalog.ldc.upenn.edu/LDC2013T19}}~\cite{pradhan2013towards}, MultiCoNER\footnote{\href{https://registry.opendata.aws/multiconer/}{https://registry.opendata.aws/multiconer/}}\cite{malmasi-etal-2022-multiconer}, MRPC\footnote{\href{https://www.microsoft.com/en-us/download/details.aspx?id=52398}{https://www.microsoft.com/en-us/download/details.aspx?id=52398}}\cite{dolan2005automatically} and the Quora Question Pairs (QQP) \footnote{\href{https://quoradata.quora.com/First-Quora-Dataset-Release-Question-Pairs}{https://quoradata.quora.com/First-Quora-Dataset-Release-Question-Pairs}}, SQuAD\footnote{\href{https://rajpurkar.github.io/SQuAD-explorer/}{https://rajpurkar.github.io/SQuAD-explorer}}~\cite{rajpurkar2016squad}, NewsQA\footnote{\href{https://www.microsoft.com/en-us/research/project/newsqa-dataset/download/}{https://www.microsoft.com/en-us/research/project/newsqa-dataset/download/}}~\cite{trischler-etal-2017-newsqa}, SNLI\footnote{\href{https://nlp.stanford.edu/projects/snli/}{https://nlp.stanford.edu/projects/snli/}}~\cite{bowman-etal-2015-large} and MNLI\footnote{\href{https://cims.nyu.edu/~sbowman/multinli//}{https://cims.nyu.edu/~sbowman/multinli/}}~\cite{williams-etal-2018-broad}. All the datasets have been released under various licenses for research purposes.

\vspace{1mm}

{\noindent \textbf{Potential Risks:}} Generative models learn from vast amounts of textual data, including biased or prejudiced content present on the internet. As a result, there is a risk of bias amplification, where the models unintentionally perpetuate or reinforce existing biases. Also, generative models can generate highly coherent and contextually plausible text, raising concerns regarding the potential for generating misinformation or disinformation. 
\vspace{1mm}

\begin{table*}[]
\small
\centering
\setlength{\tabcolsep}{1.8pt}
{\renewcommand{\arraystretch}{1.5}%
\begin{tabular}{p{0.4\linewidth}|p{0.15\linewidth}|p{0.15\linewidth}|p{0.3 \linewidth}}
\toprule
\multicolumn{1}{c|}{\multirow{1}{*}{\textbf{Raw document}}} &  \multicolumn{1}{c|}{\textbf{1st-step summary}} &  \multicolumn{1}{c|}{\textbf{2nd-step abstract}}                     & \multicolumn{1}{c}{\textbf{Naive Summary}}\\ \midrule 

Health authorities in New Zealand said that about 200 passengers on the Dawn Princess ship became infected with the norovirus. The ship was scheduled to leave for Australia on Monday. The last time there was a norovirus outbreak on the ship was back in 2012. According to Yahoo, health officials conducted a series of tests, and they confirmed that the illness was norovirus, but the outbreak does seem to be going away. The norovirus usually lasts for one to three days, and those infected may experience stomach pains, vomiting, diarrhea and nausea. Princess Cruises released a statement saying that those who were infected were isolated in their cabins. They remained there until they were considered not contagious. The statement continued to say that crew members disinfected door handles, railings, elevator buttons and so forth. The cruise operator also said that passengers were encouraged to wash their hands properly and that they should use sanitizing gels. About a month ago, another cruise ship, the Crown Princess, had an outbreak of the norovirus. In that incident more than 150 crew members and passengers came down with the norovirus. That ship was also operated by Princess Cruises. & Approximately 200 passengers on the Dawn Princess ship were infected with the norovirus, prompting health authorities to conduct tests and confirm the outbreak, isolate infected individuals, and implement disinfection measures to contain the spread of the virus. & A sudden outbreak of a viral infection, norovirus, affects a significant number of passengers on a ship, prompting immediate health measures to contain its spread. & Approximately 200 passengers on the Dawn Princess ship were infected with the norovirus, according to health authorities in New Zealand. The ship was set to depart for Australia on Monday. This is not the first time the ship has experienced a norovirus outbreak, as a similar incident occurred in 2012. Health officials conducted tests and confirmed the illness was norovirus, but the outbreak appears to be subsiding. Those infected may experience stomach pains, vomiting, diarrhea, and nausea for one to three days. Princess Cruises stated that infected passengers were isolated in their cabins until they were no longer contagious. Crew members also disinfected surfaces and encouraged passengers to wash their hands properly and use sanitizing gels. Another Princess Cruises ship, the Crown Princess, experienced a norovirus outbreak last month, infecting over 150 crew members and passengers.\\ \midrule

After the martyrdom of St. Boniface, Vergilius was made Bishop of Salzburg (766 or 767) and laboured successfully for the upbuilding of his diocese as well as for the spread of the Faith in neighbouring heathen countries, especially in Carinthia. He died at Salzburg, 27 November, 789. In 1233 he was canonized by Gregory IX. His doctrine that the earth is a sphere was derived from the teaching of ancient geographers, and his belief in the existence of the antipodes was probably influenced by the accounts which the ancient Irish voyagers gave of their journeys. This, at least, is the opinion of Rettberg ("Kirchengesch. Deutschlands", II, 236). & Vergilius, Bishop of Salzburg, spread the faith and built his diocese, and his teachings on the earth's shape were influenced by ancient geographers and Irish voyagers. & A religious leader's efforts to spread the faith and build his diocese, accompanied by teachings on the earth's shape inspired by ancient sources and travelers' accounts. & Vergilius was appointed Bishop of Salzburg in 766 or 767 after the martyrdom of St. Boniface. He worked to strengthen his diocese and spread Christianity to nearby pagan countries, particularly Carinthia. He died on November 27, 789, and was canonized by Gregory IX in 1233. Vergilius believed in the spherical shape of the earth and the existence of antipodes, ideas influenced by ancient geographers and Irish voyagers' accounts of their journeys.\\ \midrule

A blind man in his 60s is searching for the young man who pulled him back from an approaching train and saved his life last Thursday, Nov. 9. Mike Wyatt stood at the stairs to the Peoria train station in Aurora, Colorado, on Monday, looking for the young man who saved his life days earlier. Wyatt was heading back home after visiting friends in Longmont, and was about to cross the tracks to transfer to another train, unaware that a train was approaching. Seconds before the train pulled up, a man can be seen on security camera footage pulling him back with both arms. “I am so stoked right now thinking that guy is going to come down that ramp [from the platform],” Wyatt told 9 News. In the video by 9 News, he can be seen talking to passersby in hopes that they have some info about the man. Although he hasn’t yet found him, 9 News did find a woman who saw the incident. She said she felt inspired by it. “People are good,” said Miranda, one of the witnesses. 9News writes that Wyatt is thankful because the man has made it possible for Wyatt to be part of his grandchildren’s lives as they grow up. Wyatt told 9 News he will come to the station one more day to look for the man. “I will be always remembering this man and his kindness,” he said. & A blind man in his 60s is searching for a young man who saved his life by pulling him back from an approaching train, and is hoping to find him to express his gratitude. & A blind man is searching for a young hero who saved his life from a train, hoping to express gratitude. & A blind man named Mike Wyatt is searching for a young man who saved his life by pulling him back from an approaching train in Aurora, Colorado. Wyatt was unaware of the train's approach and was about to cross the tracks when the young man intervened. Despite not being able to find the man yet, Wyatt is grateful for his kindness and plans to continue searching for him. A witness named Miranda was inspired by the act of kindness and Wyatt is thankful for the opportunity to be part of his grandchildren's lives.\\

\bottomrule           
\end{tabular}
}
\caption{\small Example instances from $\mathcal{D}_{ab}$. The 1st-step and the 2nd-step abstract denote the outputs of both prompts employed in constructing $\mathcal{D}_{ab}$. Additionally, we compare the abstract with a naive summary generated using the same LLM to show the difference between the two.}
\label{tab:examples_1}
\end{table*}

\begin{table*}[]
\small
\centering
\setlength{\tabcolsep}{1.8pt}
{\renewcommand{\arraystretch}{1.5}%
\begin{tabular}{p{0.4\linewidth}|p{0.15\linewidth}|p{0.15\linewidth}|p{0.3 \linewidth}}
\toprule
\multicolumn{1}{c|}{\multirow{1}{*}{\textbf{Raw document}}} &  \multicolumn{1}{c|}{\textbf{1st-step summary}} &  \multicolumn{1}{c|}{\textbf{2nd-step abstract}}                     & \multicolumn{1}{c}{\textbf{Naive Summary}}\\ \midrule

After Kevin’s horrific leg injury, everyone is asking: will he ever play basketball again? Dr. Glashow, co-chief of sports medicine at Mount Sinai Medical Center, tells HollywoodLife.com exclusively if Kevin will ever be back on the court. Read on for details! After Louisville sophomore Kevin Ware broke his right leg during a game against Duke University, the basketball player was rushed to the hospital and underwent emergency surgery. After a successful surgery, Kevin is now focused on healing, but what does the future for his career hold? Kevin Ware’s Basketball Career: Will He Ever Play Again? Dr. Glashow, the co-chief of sports medicine at Mount Sinai Medical Center, tells HollywoodLife.com exclusively that not only will Kevin play basketball again, but he could be healed in time for next season! Although Kevin will be able to play basketball again, he has a long road to recovery ahead of him before he’ll be back on the court. & Dr. Glashow, a sports medicine expert, predicts that Kevin Ware will be able to play basketball again, potentially as soon as next season, following a successful surgery and a long road of physical therapy and recovery. & A sports medicine expert predicts a basketball player's recovery and potential return to the game following surgery and physical therapy. & The article discusses the possibility of Kevin Ware, a Louisville sophomore basketball player, returning to the court after a horrific leg injury. Dr. Glashow, a sports medicine specialist, tells HollywoodLife.com that Kevin will likely play basketball again, possibly even next season, but he has a long road to recovery ahead of him. The article also mentions that Kevin will undergo physical therapy as part of his recovery process. The public is then asked to vote on whether they think Kevin will play basketball again.\\ \midrule

Karl Kispert, principal of cyber and information security, has more than 28 years of experience in selling, managing and delivering information risk management, internal audit, regulatory and compliance programs, and information security and technology risk management. A former chief information security officer, Kispert has helped design and implement cybersecurity programs for many firms, according to the firm. “By adding this new service line, and bringing someone with Karl’s expertise to the firm, we can service yet another important aspect of our clients’ and prospects’ businesses, ensuring their continued success,” CEO Louis Grassi said in a written statement. Services will include full security programs, compliance, third party vendor risk assessment, threat management, and managed security services. & The article discusses the addition of a new cybersecurity service line at a company, led by experienced principal Karl Kispert, which will offer various services including full security programs, compliance, third-party vendor risk assessment and threat management to help clients ensure their continued success. & A company expands its services to include cybersecurity, led by an experienced principal, offering various services to help clients protect their success. & The article announces that a company has hired Karl Kispert, a seasoned professional with over 28 years of experience in cybersecurity and information risk management, as the principal of cyber and information security. Kispert has helped design and implement cybersecurity programs for many firms and will now lead the company's new cybersecurity service line. The service line will offer a range of services, including full security programs, compliance, third-party vendor risk assessment, threat management, and managed security services, to help clients ensure their continued success. The CEO, Louis Grassi, believes that Kispert's expertise will enable the company to provide an additional important service to its clients and prospects. \\ \midrule

A version of this article appeared in the print edition of The Daily Star on July 12, 2018, on page 16. Assembled by Palestinian artist Said Baalbaki, the exhibition presents 50 of Abbo's sketches, etchings and object, as well as texts – all taken from Baalbaki's personal collection of work from, and publications on, the elusive sculptor. Baalbaki has no work of his own in this show. Born in Safad, northern Palestine, around 1888-1890, Abbo was a farmer and fisherman who later took up carpentry and stone masonry. It was not until he moved to Berlin in 1911 to study sculpture that his artistic career took shape. Baalbaki has spent the last four years finding, and determining attribution for, 70 artworks and 100 texts mentioning the artist. Baalbaki says many of Abbo's sculptures didn't survive World War I. He believes those that were originally exhibited in museums in Germany were taken during the war and melted down to make weapons. Alongside Abbo's artworks are vintage photos of places he'd been or postcards from that time period, which Baalbaki says is to give an overall look at the life he led, in today's age of media and imagery. Baalbaki intends to continue his research and tracking down Abbo's works. & The article discusses an exhibition of works by the elusive sculptor Abbo, curated by Palestinian artist Said Baalbaki, featuring 50 sketches, etchings, objects, and texts from Baalbaki's personal collection, with a focus on Abbo's life and career, including his time in Berlin and the impact of World War I on his art. & An exhibition showcasing the work and life of an obscure sculptor, Abbo, curated by Said Baalbaki, featuring various artworks and personal texts. & The article discusses an exhibition of works by the elusive sculptor, Abbo, held at the Palestinian artist Said Baalbaki's studio. The exhibition features 50 of Abbo's sketches, etchings, and objects, as well as texts from Baalbaki's personal collection. Baalbaki has spent four years researching and tracking down Abbo's works, and believes that many of his sculptures did not survive World War I. The exhibition also includes vintage photos and postcards to provide context on Abbo's life. Baalbaki plans to continue his research and tracking down more of Abbo's works. \\

\bottomrule           
\end{tabular}
}
\caption{\small Example instances from $\mathcal{D}_{ab}$. The 1st-step and the 2nd-step abstract denote the outputs of both prompts employed in constructing $\mathcal{D}_{ab}$. Additionally, we also compare the abstract with a naive summary generated using the same LLM to show the difference between the both.}
\label{tab:examples_2}
\end{table*}

\section{Augmentation Examples}
\label{sec:aug_example}

\begin{figure*}
    \includegraphics[width=1\textwidth]{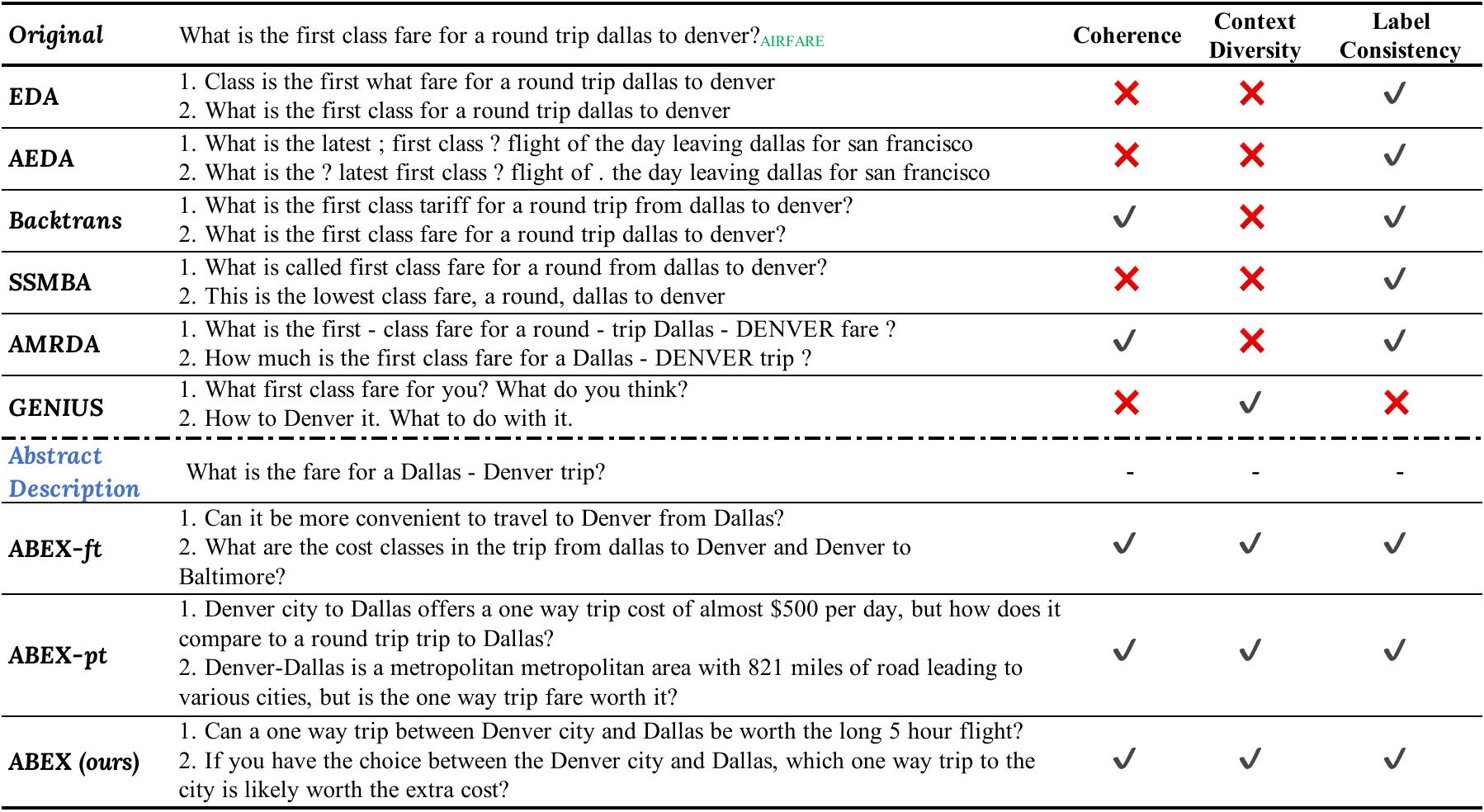}
    \caption{\small{Augmentation examples on the ATIS dataset. All generations are produced in a low-resource setting (500 training examples).}}
    \label{fig:atis}
\end{figure*}

\begin{figure*}
    \includegraphics[width=1\textwidth]{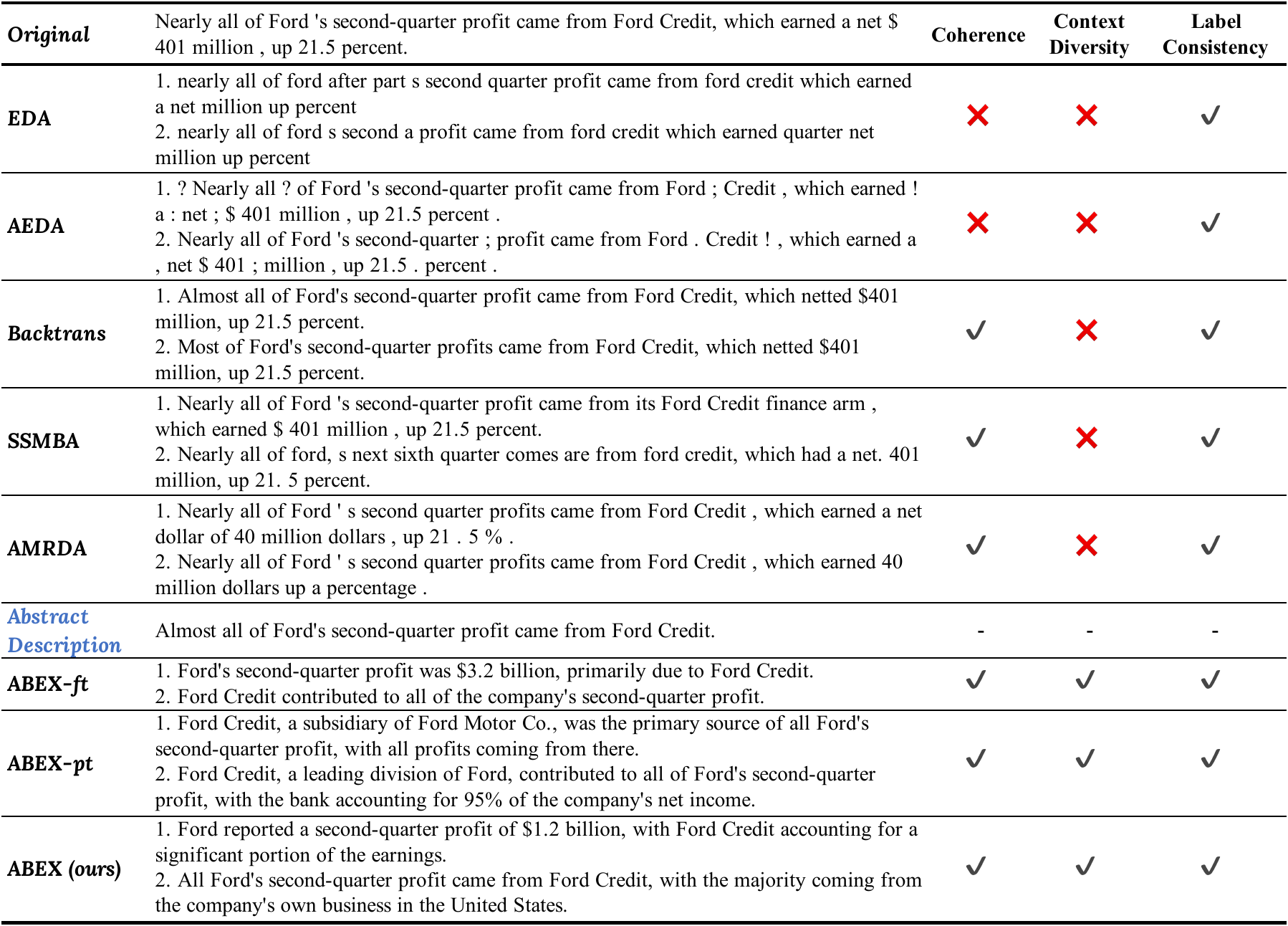}
    \caption{\small{Augmentation examples on the MRPC dataset. All generations are produced in a low-resource setting (500 training examples).}}
    \label{fig:mrpc}
\end{figure*}

\begin{figure*}
    \includegraphics[width=1\textwidth]{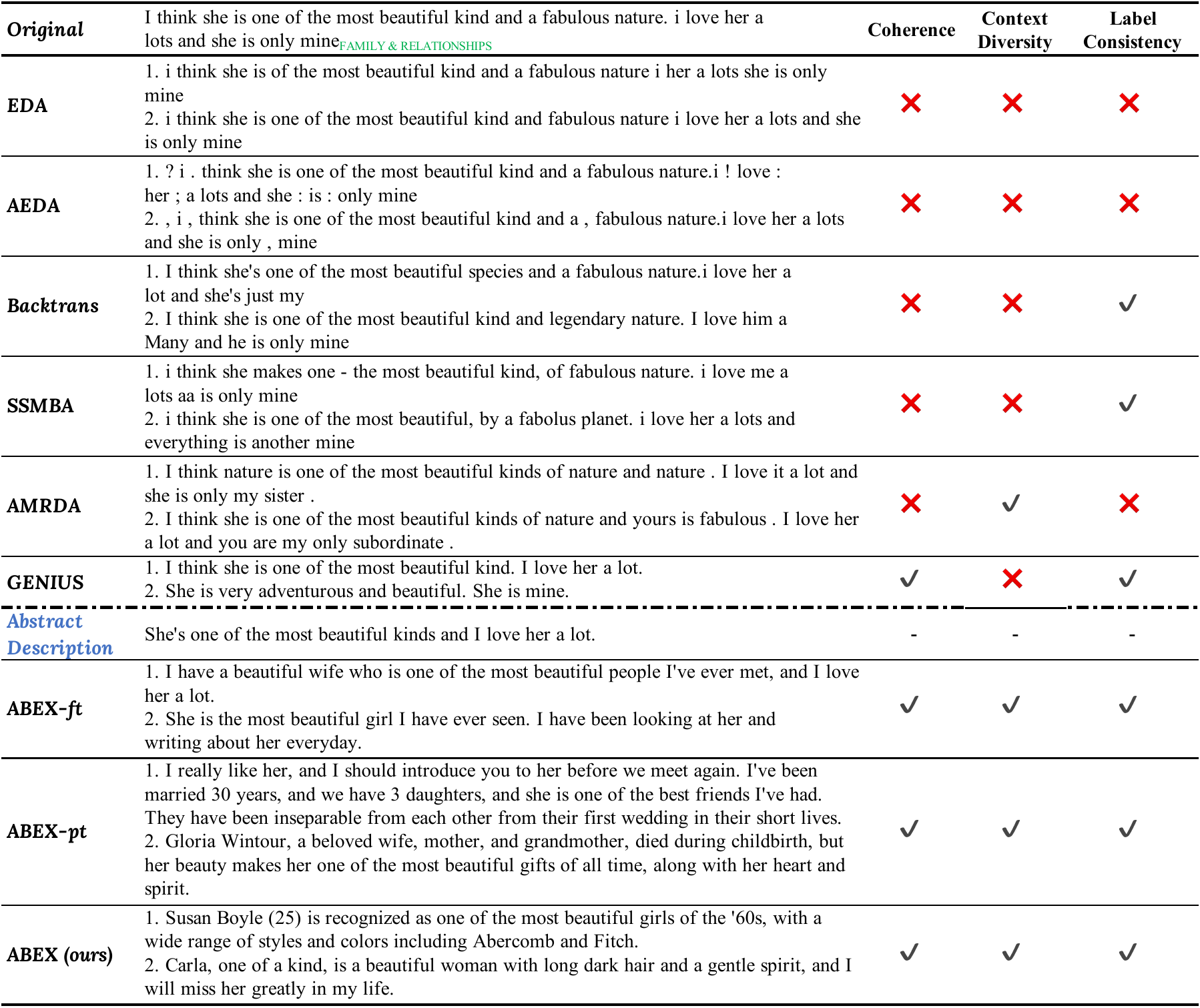}
    \caption{\small{Augmentation examples on the Yahoo dataset. All generations are produced in a low-resource setting (500 training examples).}}
    \label{fig:yahoo}
\end{figure*}

 Figure \ref{fig:atis}, Figure \ref{fig:mrpc} and Figure \ref{fig:yahoo} compare augmentations generated by ABEX with all our baselines. The figures show generations from the ATIS \cite{CNTK_2023}, Yahoo \cite{zhang2015character} and MRPC \cite{dolan2005automatically} datasets. In addition, we assess the augmentations on their coherence, ability to include diverse contexts and maintain label consistency. Notably, all baselines demonstrate the ability to generate augmentations with label consistency. However, they fall short of introducing new contextual information within the sentences. Conversely, augmentations generated by AMR-DA and Backtrans. consistently exhibit coherence, while those produced by AEDA and SSMBA often lack coherence. The generations from ABEX excel in all three evaluated areas.



\begin{figure*}
    \includegraphics[width=1\textwidth]{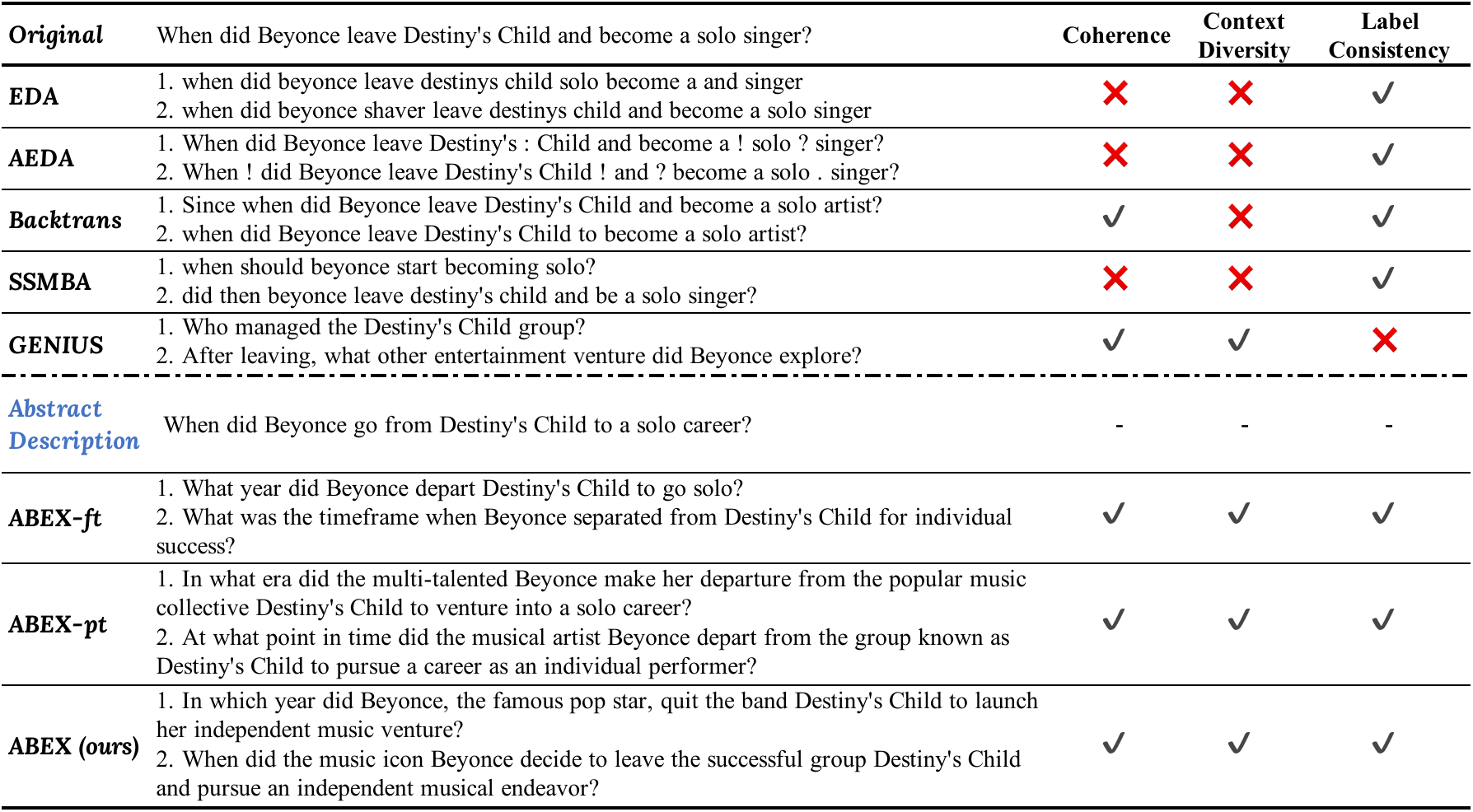}
    \caption{\small{Augmentation examples on the SQuAD dataset. All generations are produced in a low-resource setting (500 training examples).}}
    \label{fig:qa-abex}
\end{figure*}

\end{document}